\documentclass[10pt,journal,compsoc]{IEEEtran}
\usepackage[american]{babel}

\usepackage{graphicx}
\usepackage{amsmath}
\usepackage{amssymb}
\usepackage{mathrsfs}

\usepackage{bm}
\usepackage{booktabs}
\usepackage[ruled,linesnumbered]{algorithm2e}
\usepackage{subcaption}
\usepackage{lipsum}
\usepackage{colortbl}
\definecolor{mygray}{gray}{.8}

\usepackage[pagebackref=false,breaklinks=false,letterpaper=false,colorlinks,bookmarks=false]{hyperref}

\newcommand{\secref}[1]{$\S$ \ref{#1}}

\def\ie{\emph{i.e.}}
\def\eg{\emph{e.g.}}

\def\etal{{\em et al.~}}

\usepackage{wrapfig}
\usepackage{tikz,pgfplots}

\usepackage{ragged2e}
\renewcommand{\raggedright}{\leftskip=0pt \rightskip=0pt plus 0cm}

\ifCLASSOPTIONcompsoc
  \usepackage[nocompress]{cite}
\else
  \usepackage{cite}
\fi

\begin{document}

\title{Consistency and Diversity induced Human Motion Segmentation }

 \author{Tao Zhou,~\IEEEmembership{Member,~IEEE,}
 	Huazhu Fu,~\IEEEmembership{Senior~Member,~IEEE,}
 	Chen~Gong,
 	Ling Shao,~\IEEEmembership{Fellow,~IEEE,}\\
 	Fatih Porikli,~\IEEEmembership{Fellow,~IEEE},
 	Haibin~Ling,
 	Jianbing Shen,~\IEEEmembership{Senior~Member,~IEEE}\\

\IEEEcompsocitemizethanks{
\IEEEcompsocthanksitem T. Zhou and C.~Gong are with the PCA lab, the Key Laboratory of Intelligent Perception and Systems for High-Dimensional Information of Ministry of Education, 
School of Computer Science and Engineering, Nanjing University of Science and Technology, Nanjing 210094, China.
(e-mails: taozhou.dreams@gmail.com, chen.gong@njust.edu.cn).
\IEEEcompsocthanksitem H.~Fu is with Institute of High Performance Computing, Agency for Science, Technology and Research, Singapore (e-mail: hzfu@ieee.org).
\IEEEcompsocthanksitem L.~Shao is with National Center for Artificial Intelligence (NCAI), Saudi Data and Artificial Intelligence Authority (SDAIA), Riyadh, Saudi Arabia (e-mail: ling.shao@ieee.org).

\IEEEcompsocthanksitem F.~Porikli is with the Research School of Engineering, the Australian
National University (email: fatih.porikli@anu.edu.au).
\IEEEcompsocthanksitem H. Ling is with the Department of Computer Science, Stony Brook
University, Strony Brook, NY, USA (Email: hling@cs.stonybrook.edu).
\IEEEcompsocthanksitem J.~Shen is with the State Key Laboratory of Internet of Things for Smart City, Department of Computer and Information Science, University (e-mail: shenjianbingcg@gmail.com).
%\IEEEcompsocthanksitem A preliminary version of this work has appeared in CVPR 2020~\cite{zhoucvpr}.
\IEEEcompsocthanksitem Corresponding author: Jianbing Shen.
}
}

\markboth{IEEE Transactions on Pattern Analysis and Machine Intelligence}%
{Shell \MakeLowercase{\textit{et al.}}: Bare Demo of IEEEtran.cls for Computer Society Journals}

\IEEEtitleabstractindextext{
\begin{abstract}

\raggedright{Subspace clustering is a classical technique that has been widely used for human motion segmentation and other related tasks. However, existing segmentation methods often cluster data without guidance from prior knowledge, resulting in unsatisfactory segmentation results. To this end, we propose a novel \textbf{C}onsistency and \textbf{D}iversity induced human \textbf{M}otion \textbf{S}egmentation (CDMS) algorithm. Specifically, our model factorizes the source and target data into distinct multi-layer feature spaces, in which transfer subspace learning is conducted on different layers to capture multi-level information. A multi-mutual consistency learning strategy is carried out to reduce the domain gap between the source and target data. In this way, the domain-specific knowledge and domain-invariant properties can be explored simultaneously. Besides, a novel constraint based on the Hilbert Schmidt Independence Criterion (HSIC) is introduced to ensure the diversity of multi-level subspace representations, which enables the complementarity of multi-level representations to be explored to boost the transfer learning performance. Moreover, to preserve the temporal correlations, an enhanced graph regularizer is imposed on the learned representation coefficients and the multi-level representations of the source data. The proposed model can be efficiently solved using the Alternating Direction Method of Multipliers (ADMM) algorithm. Extensive experimental results on public human motion datasets demonstrate the effectiveness of our method against several state-of-the-art approaches}.

\end{abstract}

}

\maketitle

\IEEEdisplaynontitleabstractindextext

\IEEEpeerreviewmaketitle

\IEEEraisesectionheading{\section{Introduction}\label{sec:introduction}}

\IEEEPARstart{H}{uman} motion segmentation has received widespread interest in industry and research communities due to its extensive applications in video retrieval, virtual reality, and smart surveillance for user interfaces and human action analysis \cite{poppe2007vision,zhou2012hierarchical,keuper2018motion}. The main goal of human motion segmentation is to partition temporal data sequences that depict human activities and actions into a set of non-overlapping and internally temporal segments. More importantly, it is often used as a preprocessing step before motion-related analytical tasks. Currently, there are several significant challenges for human motion segmentation, including the lack of well-defined activity fragments, and the ambiguity in motion primitives caused by temporal variability among different actions \cite{lin2016movement}.

\begin{figure}[t]
	\begin{centering}
		\includegraphics[width=0.48\textwidth]{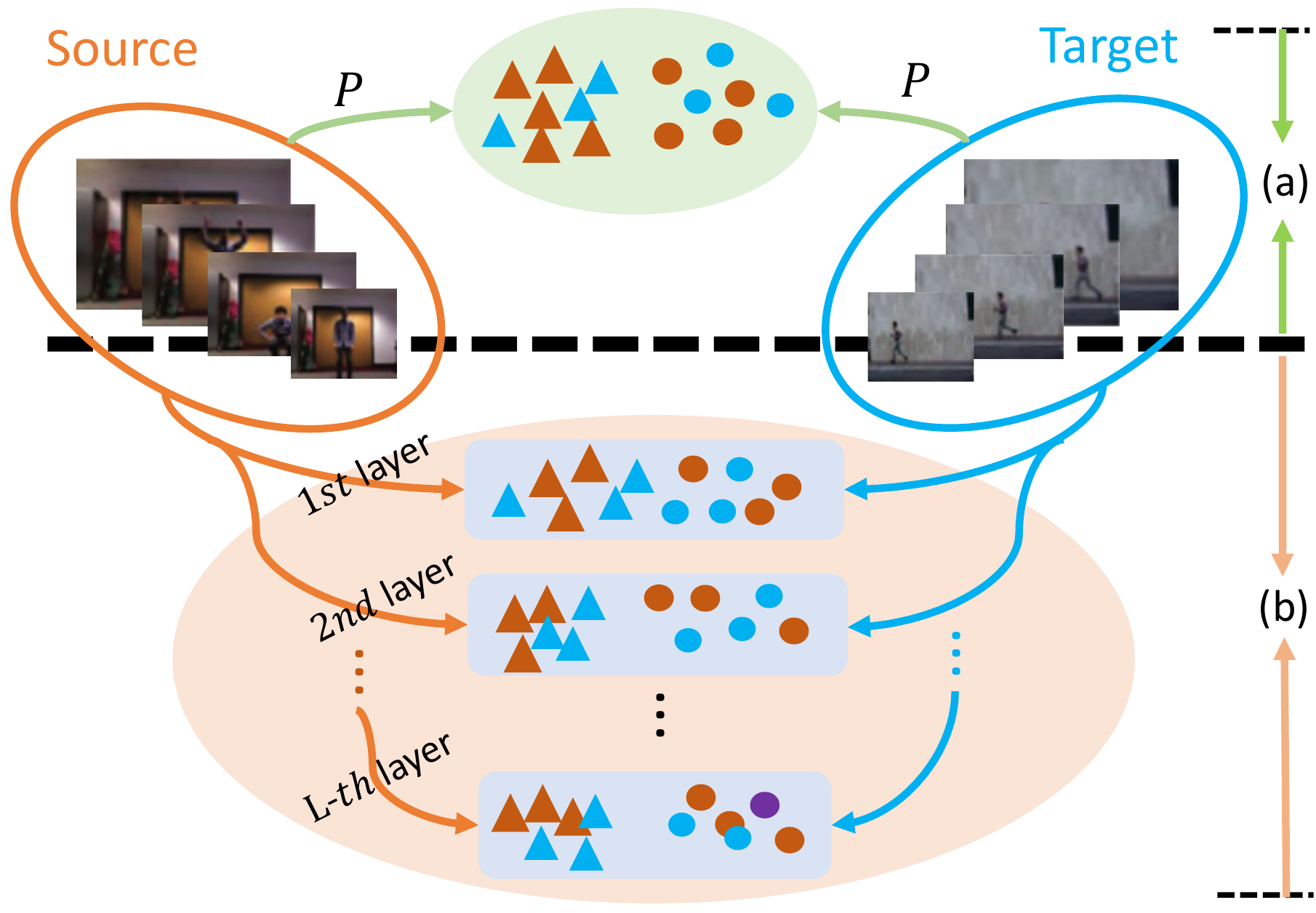}
		\caption{\footnotesize Comparison of different transfer subspace learning algorithms: (a) low-dimensional feature reconstruction based transfer subspace learning, and (b) multi-level feature reconstruction based transfer subspace learning. Different shapes denote the data points from the source or target domain, and the two colors denote different classes.}\vspace{-0.45cm}
		\label{fig01}
	\end{centering}
\end{figure}%

Human motion data contains temporal information, which is critical for motion segmentation. However, due to the high-dimensional structure of visual representations and complexity of temporal correlations, it is still challenging to capture such discriminative temporal information from time-series data \cite{keogh2003need}. Several algorithms have been proposed to address this challenge, including model-based \cite{xiong2002mixtures}, temporal proximity-based \cite{keogh2003need}, and representation-based \cite{li2015temporal}. Moreover, since temporal data occurs in the form of consecutive frame samples, human motion segmentation can be treated as an unsupervised clustering task \cite{zhou2008aligned}. Among clustering-based motion segmentation methods, subspace clustering-based approaches \cite{xia2017human,li2015temporal} have attracted increasing attention and obtained promising segmentation performance.

Subspace clustering is a popular technique to partition a given data set into different groups, based on the assumption that data points are drawn from multiple subspaces corresponding to different classes \cite{cao2015constrained,ng2002spectral}. In general, current subspace clustering approaches can be categorized into four types, including algebraic approaches \cite{costeira1998multibody}, iterative methods \cite{lu2006combined}, statistical methods, and spectral clustering methods \cite{ng2002spectral}. Over the last few decades, several representative subspace clustering methods \cite{elhamifar2013sparse,hu2014smooth,liu2013robust,lu2012robust} have been proposed to learn distinct and low-dimensional data representations, in which the learned representations can be fed to classic clustering algorithms (\eg, spectral clustering \cite{ng2002spectral}). However, these unsupervised subspace learning approaches often obtain unsatisfactory results due to the lack of some prior knowledge. Since labeled data from related tasks are often easy to obtain, transfer learning provides an effective solution to borrow knowledge from the related source data to improve the performance of target tasks \cite{cui2018large,zamir2018taskonomy}. For the human motion segmentation task, transfer subspace learning-based approaches \cite{wang2018learning,wang2018low} have been developed and demonstrated improved performance.

Although effectiveness has been achieved in transfer subspace learning-based human motion segmentation, several problems remain for existing methods. \emph{First}, subspace clustering-based approaches tend to reconstruct data points (\eg, a self-representation strategy) using low-dimensional feature representations (as shown in Fig.~\ref{fig01}), however, few approaches conduct transfer subspace learning in multi-level feature spaces to simultaneously capture low- and high-level information. \emph{Second}, transfer subspace learning forces the data distributions of two domains to be similar. To achieve this, one widely-used strategy is to project the original features of source and target data into a common low-dimensional feature space. Obviously, this strategy captures domain-invariant properties while ignoring some potentially useful domain-specific knowledge. Since both aspects are equally essential in transfer learning, it is important to balance them for boosting the model performance.

\begin{figure*}[t]
	\begin{center}
		\includegraphics[width=0.98\textwidth]{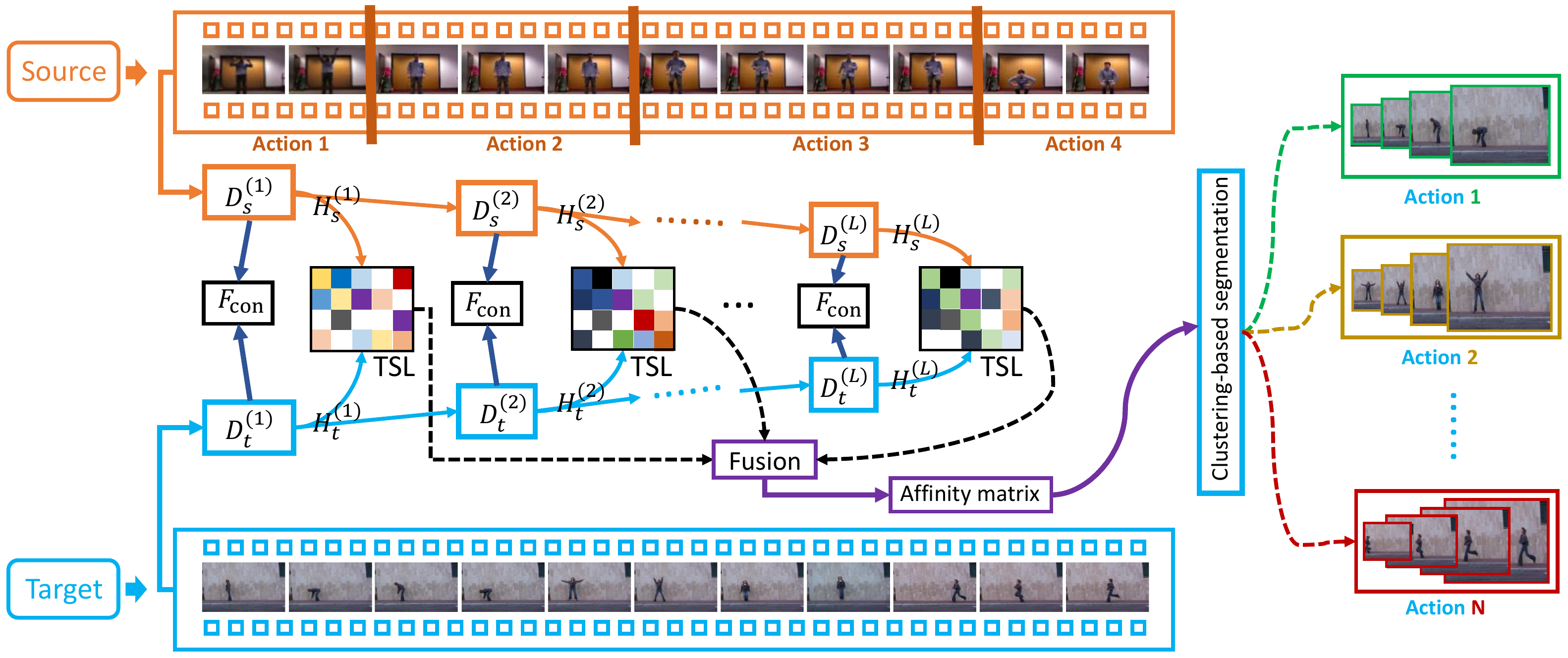}
	\end{center}\vspace{-0.25cm}
	\caption{\footnotesize Overview of multi-level transfer subspace learning framework for human motion segmentation. Our model first factorizes the source and target data into multi-layer implicit feature spaces using a deep NMF model, in which multi-level transfer subspace learning (TSL) is carried out in different spaces (\emph{i.e.,} layers) to capture multi-level information. Then, a multi-mutual consistency learning strategy is presented to reduce the difference in feature distribution between the two domains. After that, we construct a novel affinity matrix by fusing multi-level representation coefficients. Finally, the Normalized Cuts algorithm can be applied to the learned affinity matrix to obtain the segmentation (clustering) results. }\vspace{-0.25cm}
	\label{fig02}
\end{figure*}

To this end, we propose a novel \textbf{C}onsistency and \textbf{D}iversity induced human \textbf{M}otion \textbf{S}egmentation (CDMS) algorithm, which incorporates transfer learning and multi-level subspace clustering in a unified framework, to enhance human motion segmentation (as shown in Fig.~\ref{fig02}). Specifically, we first factorize the original features extracted from the source and target data into implicit multi-layer feature spaces by using a Non-negative Matrix Factorization (NMF), in which we conduct transfer subspace learning in multi-level feature spaces. A multi-mutual consistency learning strategy is proposed to reduce the distribution gap between the two domains. Moreover, we propose a novel constraint based on the Hilbert Schmidt Independence Criterion (HSIC) to enhance the diversity of the learned representation coefficients. Further, an enhanced graph regularizer is imposed on the learned representation coefficients and feature representations of the source data set to preserve the temporal correlations. Finally, we show that our model can be efficiently solved using the Alternating Direction Method of Multipliers (ADMM) algorithm. Experimental results on multiple benchmark datasets demonstrate the superiority of our model over other state-of-the-art approaches.

In summary, the key \textbf{contributions} are as follows:
\begin{itemize}{\setlength{\parsep}{-0.25ex}}
	
	\item We present a novel human motion segmentation algorithm, which integrates multi-level subspace learning and transfer learning into a unified framework. Our model aims to transfer knowledge from related source data to boost the performance of target data in the human motion segmentation task. To the best of our knowledge, this work is the first to develop multi-level subspace learning for human motion segmentation.
	
	\item  A deep NMF structure is built to capture the hidden information by leveraging the benefits of strong interpretability from the NMF model. Through this deep NMF structure, we obtain multi-level non-negative representations based on different dimensionalities of the dictionary atoms in multi-layer feature spaces, which reduce the distribution difference between the source and target domains.

	\item Our model explores domain-invariant properties by using a multi-mutual consistency learning strategy while preserving domain-specific knowledge. The main advantage of our approach is to automatically balance the two aspects for enhancing the ability of transfer subspace learning.
	
	\item We propose a novel constraint term to ensure that the learned multi-level subspace representation coefficients are diverse, which can help explore the complementary information from multi-level feature spaces to boost the segmentation performance.

\end{itemize}

This paper significantly extends our previous work in the conference paper \cite{zhoucvpr}, with multi-fold aspects. First, we reformulate a new multi-level transfer subspace learning framework with three key components, \ie, multi-mutual consistency learning, diversity cross multi-level representation, and temporal correlation preservation, for human motion segmentation. Second, we propose to utilize the HSIC as a diversity constraint for explicitly encouraging the learned multi-level representations to be of sufficient diversity, which can enhance complementary information to boost transfer subspace learning performance. Third, we provide deeper insight into the proposed multi-level transfer subspace learning framework. In this regard, our model is a general approach, as both hand-crafted HOG features and deep CNN features can be fed into our model. Last but not least, more experiments and additional ablation studies are carried out to further investigate the effectiveness of the proposed model and different key components.

The rest of this paper is arranged as follows. Related works, including subspace clustering, human motion analysis, temporal data clustering, and transfer learning, are briefly introduced in Section~\ref{reworks}. We present the details of the proposed method in Section~\ref{Methodology}. We provide the experimental settings, experimental results, and model study in Section~\ref{Experiments}. Finally, we conclude this paper in Section~\ref{Conclusion}.

\section{Related Work}
\label{reworks}
There are four types of works that are most related to the proposed human motion segmentation method, including 1) subspace clustering, 2) human motion analysis, 3) temporal data clustering, and 4) transfer learning.

\subsection{Subspace Clustering}
Subspace clustering \cite{vidal2011subspace,zhang2017latent,zhou2019dual,yang2019subspace,lu2018} holds the assumption that data points can be drawn from multiple subspaces corresponding to different clustering groups. Currently, self-representation based subspace clustering is increasing attention, in which each data point is expressed using a linear combination of other data points.
For example, sparse subspace clustering (SSC) \cite{elhamifar2013sparse} aims to find the sparsest representation among the infinitely possible representations based on $\ell_1$-norm. Different from SSC, low-rank representation clustering (LRR) \cite{liu2013robust} attempts to uncover hidden structures with a low-rank representation. By introducing a graph regularizer, smooth representation clustering (SMR) \cite{hu2014smooth} investigates the grouping effect of representation-based algorithms. Moreover, there are also several deep learning-based subspace clustering approaches \cite{ji2017, pe2018tip, zhang2018pami, zhao2017multi, zhou2018cvpr}. However, these methods cannot be directly applied to motion segmentation since they do not consider the temporal correlations in successive frames.

\subsection{Human Motion Analysis}

Over the past decades, several methods have been developed for human motion analysis. For example, Zhong \etal \cite{zhong2004detecting} develop a bipartite graph co-clustering framework to segment unusual activities in video. Jenkins \etal \cite{fod2002automated} adopt the zero-velocity crossing points of the angular velocity to partition a stream of motion data into different sequences. Barbic \etal \cite{barbivc2004segmenting} develop to decompose human motion into multiple distinct actions using the probabilistic principal component analysis algorithm. Further, Beaudoin \etal \cite{beaudoin2008motion} present a new framework to automatically distill a motion-motif graph from an arbitrary collection of motion data. Moreover, several clustering-based approaches have been proposed to segment a stream of human behavior into several activities \cite{de2007temporal}.

\subsection{Temporal Data Clustering}

The goal of the temporal data clustering task is to segment data sequences into a set of non-overlapping parts. It has a wide range of applications, from facial analytics, and speech segmentation to human action recognition. To achieve this, a semi-Markov K-means clustering \cite{robards2009semi} model is often used to exploit repetitive patterns. For instance, Zhou \emph{et al.} \cite{zhou2012hierarchical} develop a K-means kernel associated with a dynamic temporal alignment framework. Temporal subspace clustering (TSC) \cite{li2015temporal} method learns expressive coding coefficients based on a non-negative dictionary learning algorithm, which also introduces a temporal Laplacian regularization term to exploit the temporal correlations. Transfer subspace segmentation (TSS) \cite{wang2018learning} borrows knowledge from relevant source data to boost the target tasks. Low-rank transfer subspace (LTS) \cite{wang2018low} method utilizes a domain-invariant projection to reduce the distribution gaps between the two domains, which constructs a graph regularizer to capture the temporal correlations. Sun \etal \cite{sun2019online} develop an online multi-task clustering framework for multi-agent human motion segmentation, which formulates a linear encoder-decoder architecture. These temporal clustering methods are all formulated as unsupervised learning frameworks, some of which adopt a self-representation strategy to achieve the motion segmentation task.

\subsection{Transfer Learning}

Transfer learning aims to borrow prior knowledge from related tasks as source data to improve the performance of target tasks. Plenty of transfer learning models \cite{gong2013connecting,ni2013subspace,shekhar2013generalized,bruzzone2009domain,li2017domain,zhang2017joint} have been developed and obtain promising performance. Among these methods, domain-invariant feature learning is a popular strategy \cite{gong2013connecting} to learn a common feature space where both the domain shift and distribution difference can be mitigated. Several works, such as dictionary learning \cite{geng2016deep,zhu2014weakly} and subspace learning \cite{shao2014generalized,xu2015discriminative}, explore the alignment of two domains. Moreover, deep learning has been introduced to integrate knowledge transfer and feature learning into a unified framework \cite{ding2018deep,ganin2014unsupervised,long2015learning,tzeng2015simultaneous,long2016deep,long2018transferable}. However, most of these methods conduct the domain alignment using high-level features from top layers, while ignoring the low-level structural information. Besides, they mainly focus on domain-invariant feature learning without preserving domain-specific knowledge.

\renewcommand\arraystretch{1.0}
\begin{table}
	\centering
	\scriptsize
	\caption{Main notations used in the proposed model.}\vspace{-0.25cm}
	\begin{tabular}{p{0.9cm}||p{6.1cm}} 
		
		\hline
	
		Notation       & Description \\
		\hline
		$\mathbf{X}_s
		$
		& Feature matrix of the source data \\
		$\mathbf{X}_t
		$      & Feature matrix of the target data \\
		$\mathbf{D}_s^{(l)}
		$      & Basis matrix of the source data in the $l$-th layer\\
		
		$\mathbf{D}_t^{(l)}$       & Basis matrix of the target data in the $l$-th layer\\
		
		$\mathbf{H}_s^{(l)}$      & Representation matrix of the source data in the $l$-th layer\\
		$\mathbf{H}_t^{(l)}$       & Representation matrix of the target data in the $l$-th layer\\
		
		$\mathbf{Z}_s^{(l)}$      & Representation coefficient of $\mathbf{X}_s$ in the $l$-th layer\\
		$\mathbf{Z}_t^{(l)}$       & Representation coefficient of $\mathbf{X}_t$ in the $l$-th layer\\
		
		$\mathbf{L}$                 &   Graph Laplace matrix\\
		
		\hline
		
		$\alpha$, $\beta$, $\gamma$       & Trade-off parameters\\

		\hline
	\end{tabular}

	\label{tab01}\vspace{-0.05cm}
\end{table}

\section{The Proposed Method}
\label{Methodology}

In this section, we first introduce the motivation (\secref{motivation}) and provide details of the proposed method (\secref{our_model}). Moreover, we describe the optimization steps for solving the proposed model (\secref{optimization}). Finally, we construct a new affinity matrix for clustering-based motion segmentation (\secref{clustering}), and provide complexity analysis (\secref{complexity}).

\subsection{Motivation}
\label{motivation}

As previously mentioned, three main challenges remain for human motion segmentation using transfer subspace learning, \ie, \textit{(1)} how to capture multi-level information to enhance the performance of transfer subspace learning; \textit{(2)} how to decrease the distribution gap between the two domains while also preserving domain-specific knowledge; and \textit{(3)} how to effectively preserve temporal correlations among motion data. To address these challenges, we propose a novel multi-level transfer subspace learning framework with three key components for human motion segmentation, and the details of each component will be provided in the following subsections.

Moreover, deep structure learning has proven its effectiveness in several machine learning and computer vision applications \cite{trigeorgis2016deep, zhao2017multi,jiang2019heterogeneous,pan2016hierarchical}. To effectively capture multi-level structural information, we adopt a multi-layer decomposition process based on the deep NMF model, which can be formulated as:
\begin{eqnarray}
\small
\begin{split}
\mathbf{X}&\approx{\mathbf{D}^{(1)}\mathbf{H}^{(1)}}\\
&\approx{\mathbf{D}^{(1)}\mathbf{D}^{(2)}\mathbf{H}^{(2)}}\\
&~~\vdots\\
&\approx{\mathbf{D}^{(1)}\mathbf{D}^{(2)}\cdots\mathbf{D}^{(l)}\cdots\mathbf{D}^{(L)}\mathbf{H}^{(L)}},\\
\end{split}
\label{eq0000-01}
\end{eqnarray}
\noindent where {\small$\mathbf{D}^{(l)}\geq{0}$} and {\small$\mathbf{H}^{(l)}\geq{0}$ ($l=1,\ldots,L$)} represent the base matrix and feature representation in the $l$-th layer, respectively. And $L$ denotes the number of layers. Through this deep NMF structure, we obtain multi-level non-negative representations based on different dimensionalities of the dictionary atoms, which can reduce the influence of the distribution gap between the two domains in a layer-wise fashion. It can be worth noting that the learned dictionary (base matrix) are not arbitrary in our NMF-based model with the non-negativity constraint.

\begin{figure}[t]
	\begin{center}
		\includegraphics[width=0.49\textwidth]{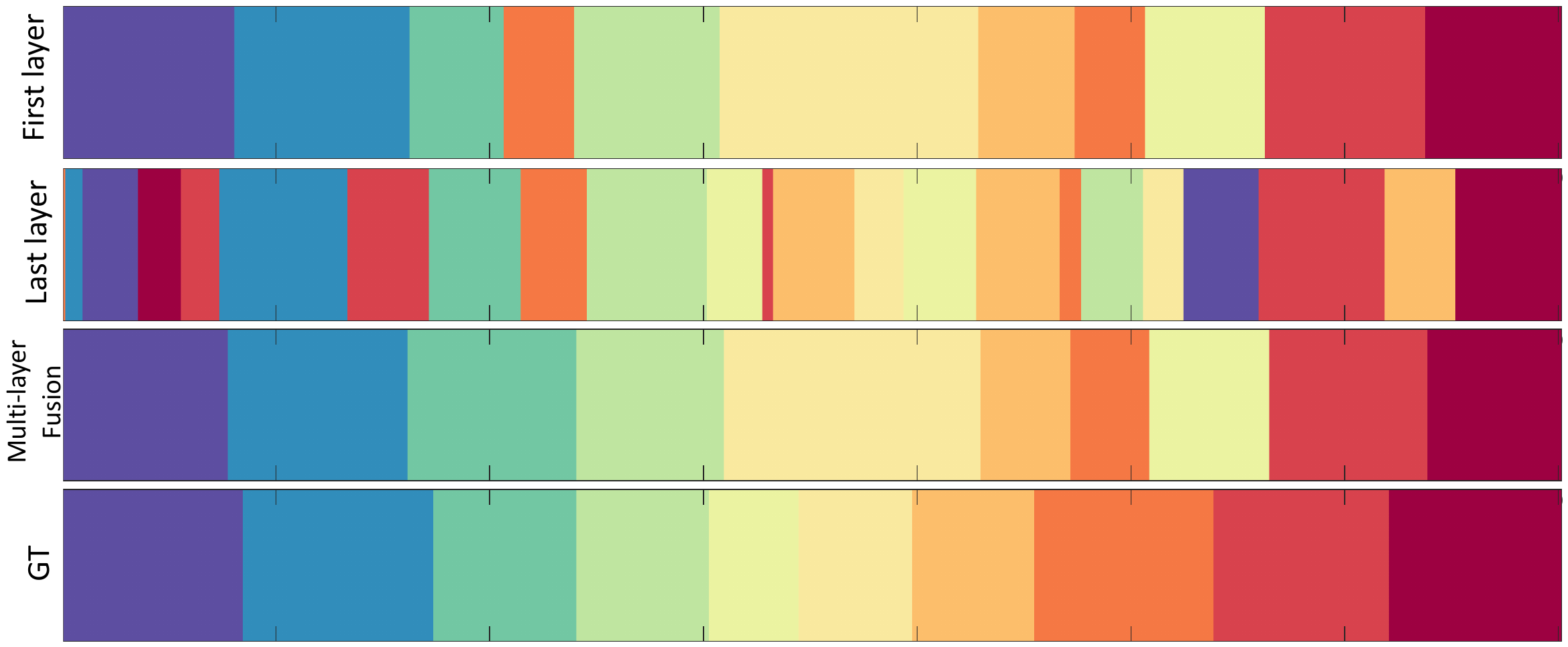}\vspace{-0.05cm}
		\caption{\footnotesize Comparison segmentation results of using different coefficient representations from the first layer, last layer, and multi-layer fusion to construct the affinity matrix on the Weiz dataset.}\vspace{-0.35cm}
		\label{fig03}
	\end{center}
\end{figure}%\vspace{-0.45cm}

%\vspace{-3pt}
\subsection{Multi-level Transfer Subspace Learning}
%\vspace{-2pt}
\label{our_model}

Existing subspace clustering or subspace clustering-based motion segmentation approaches tend to reconstruct original data points using either shallow representations (\emph{e.g.}, handcrafted features) or high-level representations (\emph{e.g.}, features from the last layer of deep networks). However, the original data often contains complex structural information and hierarchical semantics, which are difficult to extract by only using a single-layer clustering strategy. In Fig.~\ref{fig03}, we show the clustering results when constructing the affinity matrix using the coefficients from the first layer, last layer, and multi-layer fusion, respectively. As can be seen, the orange part exists in the results when using the first and last layer representation methods, which indicates that the two methods generate multiple fragments and cannot achieve meaningful or accurate segmentation. While the method using multi-level representations can accurately segment the current motion action. Motivated by these results, we propose a multi-level subspace learning strategy to effectively exploit the hierarchical semantics and structural information in a layer-wise fashion, which can be formulated as follows:
\begin{eqnarray}
\small
\begin{split}
&\mathcal{L}\left(\mathbf{X}_s,\mathbf{X}_t;\mathbf{D}_s^{(l)},\mathbf{D}_t^{(l)},\mathbf{H}_s^{(l)},\mathbf{H}_t^{(l)},\mathbf{Z}^{(l)}\right)=\\
&\|\mathbf{X}_s-\mathbf{D}_s^{(1)}\dots\mathbf{D}_s^{(L)}\mathbf{H}_s^{(L)}\|_F^2+\|\mathbf{X}_t-\mathbf{D}_t^{(1)}\dots\mathbf{D}_t^{(L)}\mathbf{H}_t^{(L)}\|_F^2\\
&~~~+\sum_{l=1}^L\|[\mathbf{H}_s^{(l)},\mathbf{H}_t^{(l)}]-\mathbf{H}_s^{(l)}\mathbf{Z}^{(l)}\|_{2,1},\\
&~~s.t.~~\mathbf{Z}^{(l)}\geq{0},\mathbf{1}^{\top}\mathbf{Z}^{(l)}=\mathbf{1}^{\top},\forall l=1,2,\dots,L,
\end{split}
\label{eq001-02}
\end{eqnarray}
\noindent where {\small$\mathbf{X}_s\in\mathbb{R}^{d\times{n_s}}$} and {\small$\mathbf{X}_t\in\mathbb{R}^{d\times{n_t}}$} denote the source and target data, respectively. {\small$\mathbf{D}_s^{(l)}\geq{0}$} and {\small$\mathbf{H}_s^{(l)}\geq{0}$ ($l=1,\ldots,L$)} denote the basis matrix and the feature representation matrix at the $l$-th layer for the source data, respectively (similar for the target data). In addition, $d$ is the original feature dimension, and $n_s$ and $n_t$ are the number of the source and target data, respectively. Here, we have {\small$\mathbf{H}_s\in\mathbb{R}^{d_l\times{n_s}}$} and {\small$\mathbf{H}_t\in\mathbb{R}^{d_l\times{n_t}}$} with $d_l$ denoting the feature dimension in the $l$-th layer, we obtain {\small$\mathbf{Z}_l\in\mathbb{R}^{n_s\times{n_s+n_t}}, l=1,2,\dots,L$}.
Moreover, {\small$\mathbf{1}$} denotes a column vector whose elements are all set to one. The first two terms aim to explore the multi-level structures in both the source and target data, and the third term is used to conduct the multi-level transfer subspace learning.

In addition, in Eq.~(\ref{eq001-02}), the non-negative constraint {\small$\mathbf{Z}^{(l)}\geq{0}$} enhances the discriminative ability of the learned representations. The constraint {\small$\mathbf{1}^{\top}\mathbf{Z}^{(l)}=\mathbf{1}^{\top}$} ensures the sum of each coefficient vector to be one, resulting in the suppression of representation coefficients from different subspaces. In Eq.~(\ref{eq001-02}), the feature representations of the source data (\emph{i.e.,} {\small$\mathbf{H}_s^{(l)}$}) are adopted as the subspace learning dictionary, which is used to reconstruct the feature representations of both the two domains (\ie, the source and target). By using this reconstruction strategy, it enables that knowledge from the related source data to be effectively transferred to the target task. Besides, $\left\|\cdot\right\|_{2,1}$ denotes the $\ell_{2,1}$-norm, which is used to constrain the columns of a matrix to be zero \cite{liu2013robust}, \emph{i.e.}, {\small$\|\textbf{E}\|_{2,1}=\sum_{j=1}^{N}\sqrt{\sum_{i=1}^{M}[\textbf{E}_{ij}]^2}$}, where {\small$\textbf{E}\in\mathbb{R}^{M\times{N}}$}. By using the {\small$\ell_{2,1}$}-norm, there is an underlying assumption that some corruptions could be sample-specific, \emph{i.e.} some data points may be corrupted while the others are clean.

\emph{\textbf{Remarks}}: Since diverse information often exists in different domains, it is difficult to transfer useful knowledge from the related source to the target by only using a single subspace. Thus, we carry out subspace learning on multi-level feature spaces to obtain multiple subspace representations and then fuse them to seek an optimal affinity matrix. Moreover, the $\ell_{2,1}$-norm has been shown to be more robust to outliers than the classic Frobenius norm \cite{liu2013robust,zhang2017latent}.

\subsubsection{Multi-mutual Consistency Learning}
To decrease the distribution gaps between the two domains as well as preserve the knowledge from different domains, we propose a multi-mutual consistency learning strategy as follows:
\begin{eqnarray}
\small
\begin{split}
&\mathcal{R}_1=\sum\nolimits_{l=1}^{L}{F}_{\textup{con}}\left(\mathbf{D}_s^{(l)},\mathbf{D}_t^{(l)}\right),\\
\end{split}
\label{eq001-03}
\end{eqnarray}
\noindent where the goal of the term {\small$F_{\textup{con}}(\cdot, \cdot)$} is to reduce the distribution gap between the two domains by penalizing the divergence of two basis matrices in different layers. In contrast, several methods directly project the original features from the source and target data into a common low-dimensional space using a domain-invariant projection matrix, which could result in a loss of an amount of domain-specific knowledge. Although there are different strategies to constrain the consistency between {\small$\mathbf{D}_s^{(l)}$} and {\small$\mathbf{D}_t^{(l)}$}, a simple but effective strategy is adopted in our model, \emph{i.e.}, {\small$F_{\textup{con}}(\mathbf{D}_s^{(l)},\mathbf{D}_t^{(l)})=\|\mathbf{D}_s^{(l)}-\mathbf{D}_t^{(l)}\|_F^2$}.

\emph{\textbf{Remarks}}: The proposed multi-mutual consistency learning strategy can reduce the distribution between source and target domains. By using the strategy, our transfer subspace learning model can transfer useful information from the source domain to benefit the segmentation while also preserving some domain-specific knowledge.

\subsubsection{Diversity across Multi-level Representation}

In the formulation of Eq.~(\ref{eq001-02}), we have presented a multi-level subspace learning strategy to effectively exploit the structural information in a layer-wise fashion. Further, to enhance the complementarity of multi-level representations, the encoded representation coefficients (\ie, {\small$\mathbf{Z}^{(l)}, l=1,2,\cdots,L$}) of different levels are encouraged to be of sufficient diversity. To achieve this, a diversity regularization term is proposed. For convenience, we define a mapping $\phi(\mathbf{x})$ from $\mathbf{x}\in\bm{\mathcal{X}}$ to kernel space $\bm{\mathcal{F}}$, where $\phi(\cdot)$ is used to map original features into kernel space. In this kernel space, the inner product between vectors is given by a kernel function  $k_1(\mathbf{x}_i,\mathbf{x}_j)=\langle \phi(\mathbf{x}_i),\phi(\mathbf{x}_j) \rangle$. Let $\bm{\mathcal{G}}$ be a second kernel space on $\bm{\mathcal{Y}}$, and we define a mapping function $\phi(\mathbf{y})$ from $\mathbf{y}\in\bm{\mathcal{Y}}$ to $\bm{\mathcal{G}}$ with $k_2(\mathbf{y}_i,\mathbf{y}_j)=\langle \phi(\mathbf{y}_i),\phi(\mathbf{y}_j) \rangle$. Inspired by \cite{gretton2005measuring}, we adopt an empirical version of HSIC to enhance the diversity of representation coefficients in the proposed model, which is given below.

\textbf{Definition 1.} Given a series of $n$ independent observations collected from $p_{xy}$, \ie, $\mathbf{Z}=\{(\mathbf{x}_1,\mathbf{y}_1),\ldots,(\mathbf{x}_n,\mathbf{y}_n)\}\subseteq\bm{\mathcal{X}}\times\bm{\mathcal{Y}}$, an estimator of $\textup{HSIC}(\textbf{Z},\bm{\mathcal{F}},\bm{\mathcal{G}})$ is defined by
\begin{eqnarray}
\begin{split}
\textup{HSIC}(\mathbf{Z},\bm{\mathcal{F}},\bm{\mathcal{G}})=(n-1)^{-2}\textup{tr}(\mathbf{K}_1\mathbf{HK}_2\mathbf{H}),
\end{split}
\label{eq012}
\end{eqnarray}
\noindent where $k_{1,ij}:=k_1(\mathbf{x}_i,\mathbf{x}_j)$ and $k_{2,ij}:=k_2(\mathbf{y}_i,\mathbf{y}_j)$. Here, $h_{ij}:=\delta_{ij}-1/n$ centralizes the two Gram matrices (\ie, $\mathbf{K}_1$ and $\textbf{K}_2$) in the feature space, ensuring that they have zero mean. Please refer to the details of HSIC in \cite{gretton2005measuring,cao2015diversity}.

Thus, to explore more complementary information from different feature spaces, we encourage the multi-level  representation coefficients {\small$\mathbf{Z}^{(l)}$} and {\small$\mathbf{Z}^{w)}$} ($w\neq{l}$) to be sufficiently diverse. The diversity regularization term can be defined by
\begin{eqnarray}
\small
\begin{split}
&\mathcal{R}_2=\sum_{l=1}^L\sum_{m\neq{l}}^L\textup{HSIC}\left(\mathbf{Z}^{(l)},\mathbf{Z}^{(m)}\right).\\
\end{split}
\label{eq001-04}
\end{eqnarray}

\emph{\textbf{Remarks}}: It is worth noting that the diversity across multi-level representations can help to explore more knowledge from source data, which could boost the performance of human motion segmentation.

%$\bullet$

\subsubsection{Temporal Correlation Preservation}

Temporal correlation is important for accurate clustering since human motion data is sequential and consecutive. Thus, it is expected to preserve the temporal information in representation coefficients {\small$\mathbf{Z}$} and feature representations {\small$\mathbf{H}$}. To achieve this, we regulate the $i$-th coefficient's neighbors {\small$[\mathbf{z}_{i-\tau/2},\cdots,\mathbf{z}_{i-1},\mathbf{z}_{i+1},\cdots,\mathbf{z}_{i+\tau/2}]$} to be close to {\small$\mathbf{z}_i$}, where $\tau$ denotes the number of sequential neighbors. Besides, to learn an effective reconstruction dictionary (\ie, {\small$\mathbf{X}_s$} in our method), we also regulate the $i$-th coefficient's neighbors {\small$[\mathbf{h}_{i-\tau/2},\cdots,\mathbf{h}_{i-1},\mathbf{h}_{i+1},\cdots,\mathbf{h}_{i+\tau/2}]$} to be close to {\small$\mathbf{h}_i$} in the source data. Thus, this enhanced graph regularizer is imposed on both the learned representation coefficients and feature representations of the source dataset. By using it, our model can effectively uncover the temporal correlations residing in both the source and target data. To achieve this, we first build a weight matrix {\small$\mathbf{S}$} \cite{li2015temporal,wang2018learning}, where each element of {\small$\mathbf{S}$} is given as:
\begin{equation}
\small
\begin{aligned}
s_{ij}=\left\{
\begin{aligned}
&1, ~\textup{if}~|i-j|\leq{\tau}, l(x_i)=l(x_j), \textup{for source data};\\
&1, ~\textup{if}~|i-j|\leq{\tau}, \textup{for target data};\\
&0,~\textup{otherwise},
\end{aligned}
\right.
\end{aligned}
\label{eq0-3}
\end{equation}
\noindent where {\small$l(x_i)$} denotes the action label of the {\small$i$}-th sample {\small$x_i$} from the source data. After that, we have the temporal correlation preservation term as follows:
\begin{equation}
\small
\begin{split}
\mathcal{R}_3=\sum_{l=1}^{L}\left(tr(\mathbf{Z}_l\mathbf{L}\mathbf{Z}_l^{\top})+tr(\mathbf{H}_s^{(l)}\mathbf{L}_s\mathbf{H}_s^{(l)^\top})\right),
\end{split}
\label{eq001-05}
\end{equation}
\noindent where $tr(\cdot)$ represents the matrix trace. $\mathbf{L}$ denotes the Laplacian matrix with $\mathbf{L}=\mathbf{C}-\mathbf{S}$, where $\mathbf{C}$ is a diagonal degree matrix with $C_{ii}=\sum_{j}S_{ij}$, and $\mathbf{L}_s$ is the corresponding part to the source data.

\textbf{Overall formulation}: Finally, we formulate the proposed CDMS model (in Eq.~\ref{eq001-02}) with three key components (in Eqs.~(\ref{eq001-03})(\ref{eq001-04})(\ref{eq001-05})) into a unified framework as follows:
\begin{equation}
\small
\begin{split}
&\min_{\Omega}\mathcal{L}+\alpha\mathcal{R}_1+\beta\mathcal{R}_2+\gamma\mathcal{R}_3\\
&=\|\mathbf{X}_s-\mathbf{D}_s^{(1)}\dots\mathbf{D}_s^{(L)}\mathbf{H}_s^{(L)}\|_F^2+\|\mathbf{X}_t-\mathbf{D}_t^{(1)}\dots\mathbf{D}_t^{(L)}\mathbf{H}_t^{(L)}\|_F^2\\
&~~~+\sum_{l=1}^L\|[\mathbf{H}_s^{(l)},\mathbf{H}_t^{(l)}]-\mathbf{H}_s^{(l)}\mathbf{Z}^{(l)}\|_{2,1}+\alpha\sum_{l=1}^{L}\|\mathbf{D}_s^{(l)}-\mathbf{D}_t^{(l)}\|_F^2\\
&~~~+\beta\sum_{l=1}^L\sum_{m\neq{l}}^L\textup{HSIC}(\mathbf{Z}^{(l)},\mathbf{Z}^{(m)})\\
&~~~+\gamma\sum_{l=1}^{L}(tr(\mathbf{Z}^{(l)}\mathbf{L}\mathbf{Z}^{(l)^{\top}})+tr(\mathbf{H}_s^{(l)}\mathbf{L}_s\mathbf{H}_s^{(l)^\top})),\\
&~~s.t.~~\mathbf{Z}^{(l)}\geq{0},\mathbf{1}^{\top}\mathbf{Z}^{(l)}=\mathbf{1}^{\top},\forall l=1,2,\dots,L,
\end{split}
\label{eq00-01}
\end{equation}
\noindent where {\small$\Omega=\{\mathbf{D}_s^{(l)}\geq{0},\mathbf{D}_t^{(l)}\geq{0},\mathbf{H}_s^{(l)}\geq{0},\mathbf{H}_t^{(l)z}\geq{0},\mathbf{Z}^{(l)},\mathbf{W}^{(l)}\}$} ({\small$l=1,2,\dots,L$}) is the variable set to be optimized, and $\alpha$, $\beta$, and $\gamma$ are trade-off parameters. For clarity, the main notations used in this paper are listed in Table~\ref{tab01}.

\subsection{Optimization}
\label{optimization}

\iffalse
Then, we have the following equivalent problem as,
\begin{eqnarray}
\begin{split}
&\min_{\Omega}~\|\mathbf{X}_s-\mathbf{D}_s^{(1)}\mathbf{D}_s^{(2)}\dots\mathbf{D}_s^{(L)}\mathbf{H}_s^{(L)}\|_F^2\\
&~~~~~~+\|\mathbf{X}_t-\mathbf{D}_t^{(1)}\mathbf{D}_t^{(2)}\dots\mathbf{D}_t^{(L)}\mathbf{H}_t^{(L)}\|_F^2\\
&~~~~~~+\alpha\sum_{l=1}^{L}\|\mathbf{D}_s^{(l)}-\mathbf{D}_t^{(l)}\|_F^2+\lambda\sum_{l=1}^L\|\mathbf{E}^{(l)}\|_{2,1}\\
&~~~~~~+\beta\sum_{l=1}^L\|\mathbf{S}-\mathbf{W}^{(l)}\mathbf{Z}^{(l)}\|_F^2+\gamma\sum_{l=1}^L\|\mathbf{W}^{(l)}\|_{*}\\
&~~~s.t.~~[\mathbf{H}_s^{(l)},\mathbf{H}_t^{(l)}]-\mathbf{H}_s^{(l)}\mathbf{Z}^{(l)}=\mathbf{E}^{(l)},\mathbf{W}^{(l)}=\mathbf{J}^{(l)},\\
&~~~~~~~~~~~\mathbf{Z}^{(l)}\geq{0},\mathbf{1}_n^{\top}\mathbf{Z}^{(l)}=\mathbf{1}_n^{\top},\forall l=1,2,\dots,L.
\end{split}
\label{eq000-01}
\end{eqnarray}
\fi

Although the proposed objective function in Eq.~(\ref{eq00-01}) is not jointly convex for all variables, we can employ the ADMM \cite{lin2011linearized} algorithm to efficiently solve it. To adopt this strategy to our problem, we introduce two auxiliary variables {\small$\textbf{J}^{(l)}$} and {\small$\textbf{E}^{(l)}$} to replace {\small${\textbf{Z}^{(l)}}$} and {\small$[\mathbf{H}_s^{(l)},\mathbf{H}_t^{(l)}]-\mathbf{H}_s^{(l)}\mathbf{Z}^{(l)}$}, respectively. Therefore, we have the following equivalent problem:
\begin{eqnarray}
\small
\begin{split}
&\mathcal{L}({\Omega})=\|\mathbf{X}_s-\mathbf{D}_s^{(1)}\dots\mathbf{D}_s^{(L)}\mathbf{H}_s^{(L)}\|_F^2\\
&~~~~~~+\|\mathbf{X}_t-\mathbf{D}_t^{(1)}\dots\mathbf{D}_t^{(L)}\mathbf{H}_t^{(L)}\|_F^2\\
&~~~~~~+\sum\nolimits_{l=1}^L\|\mathbf{E}^{(l)}\|_{2,1}+\alpha\sum\nolimits_{l=1}^{L}\|\mathbf{D}_s^{(l)}-\mathbf{D}_t^{(l)}\|_F^2\\
&~~~~~~+\beta\sum_{l=1}^L\sum_{m\neq{l}}^L\textup{HSIC}(\mathbf{Z}^{(l)},\mathbf{Z}^{(m)})\\
&~~~~~~+\gamma\sum_{l=1}^{L}(tr(\mathbf{Z}^{(l)}\mathbf{L}\mathbf{Z}^{(l)^{\top}})+tr(\mathbf{H}_s^{(l)}\mathbf{L}_s\mathbf{H}_s^{(l)^\top}))\\
&~~~~~~+\sum\nolimits_{l=1}^L\Phi\left(\bm{\Lambda}_1^{(l)},[\mathbf{H}_s^{(l)},\mathbf{H}_t^{(l)}]-\mathbf{H}_s^{(l)}\mathbf{J}^{(l)}-\mathbf{E}^{(l)}\right)\\
&~~~~~~+\sum\nolimits_{l=1}^L\Phi\left(\bm{\Lambda}_2^{(l)},\mathbf{Z}^{(l)}-\mathbf{J}^{(l)}\right),\\
&~~~s.t.~\mathbf{Z}^{(l)}\geq{0},\mathbf{1}^{\top}\mathbf{Z}^{(l)}=\mathbf{1}^{\top},\forall l=1,2,\dots,L,\\
\end{split}
\label{eq000-02}
\end{eqnarray}
\noindent where {\small$\Phi(\bm{\Lambda},\mathbf{Q})=\frac{\mu}{2}\|\mathbf{Q}\|_F^2+\langle\bm{\Lambda},\mathbf{Q}\rangle$}, with {\small$\langle \cdot,\cdot \rangle$} denoting the matrix inner product. {\small$\bm{\Lambda}_1^{(l)}$} and {\small$\bm{\Lambda}_2^{(l)}$} ({\small$l=1,2,\dots,L$}) are Lagrangian multipliers, and {\small$\mu$} is a penalty scalar. We describe the optimization steps for each subproblem below.

\textbf{{\small$\mathbf{D}_s$}-subproblem}: The optimization problem with respect to {\small$\mathbf{D}_s$} is formulated as
\begin{eqnarray}
\small
\begin{split}
&\min_{\mathbf{D}_s\geq{0}}	\|\mathbf{X}_s-\mathbf{D}_s^{(1)}\dots\mathbf{D}_s^{(L)}\mathbf{H}_s^{(L)}\|_F^2\\
&~~+\alpha\sum\nolimits_{l=1}^L\|\mathbf{D}_s^{(l)}-\mathbf{D}_t^{(l)}\|_F^2,~~\forall l=1,2,\dots,L.
\end{split}
\label{eq1-01}
\end{eqnarray}

We can update {\small$\mathbf{D}_s^{(l)}$} in each layer one by one. By taking the derivative of Eq.~(\ref{eq1-01}) with respect to {\small$\mathbf{D}_s^{(l)}$} and using the Karush-Kuhn-Tucker (KKT) condition \cite{boyd2004convex}, we have the updating rule:
\begin{equation}
\small
\begin{split}
&\mathbf{D}_s^{(l)}\leftarrow\mathbf{D}_s^{(l)}\odot\\
&\frac{{{\rm\Theta}_s^{(l-1)}}^{\top}\mathbf{X}_s{\mathbf{H}_s^{(L)}}^{\top}{{\rm\Omega}_s^{(l+1)}}^{\top}+\alpha\mathbf{D}_t^{(l)}}{{{\rm\Theta}_s^{(l-1)}}^{\top}{{\rm\Theta}_s^{(l-1)}}\mathbf{D}_s^{(l)}{\rm\Omega}_s^{(l+1)}\mathbf{H}_s^{(L)}{\mathbf{H}_s^{(L)}}^{\top}{{\rm\Omega}_s^{(l+1)}}^{\top}+\alpha\mathbf{D}_s^{(l)}},
\end{split}
\label{eq1-032}
\end{equation}
\noindent where $\odot$ denotes element-wise product, and
{\small${\rm\Theta}_s^{(l-1)}=\mathbf{D}_s^{(1)}\mathbf{D}_s^{(2)}\cdots\mathbf{D}_s^{(l-1)}$} and {\small${\rm\Omega}_s^{(l+1)}=\mathbf{D}_s^{(l+1)}\mathbf{D}_s^{(l+2)}\cdots\mathbf{D}_s^{(L)}$}.

Similarly, we have the updating rule for {\small$\mathbf{D}_t^{(l)}$} as follows
\begin{equation}
\small
\begin{split}
&\mathbf{D}_t^{(l)}\leftarrow\mathbf{D}_t^{(l)}\odot\\
&\frac{{{\rm\Theta}_t^{(l-1)}}^{\top}\mathbf{X}_t{\mathbf{H}_t^{(L)}}^{\top}{{\rm\Omega}_t^{(l+1)}}^{\top}+\alpha\mathbf{D}_s^{(l)}}{{{\rm\Theta}_t^{(l-1)}}^{\top}{{\rm\Theta}_t^{(l-1)}}\mathbf{D}_t^{(l)}{\rm\Omega}_t^{(l+1)}\mathbf{H}_t^{(L)}{\mathbf{H}_t^{(L)}}^{\top}{{\rm\Omega}_t^{(l+1)}}^{\top}+\alpha\mathbf{D}_t^{(l)}}.
\end{split}
\label{eq1-033}
\end{equation}

\textbf{{\small$\mathbf{H}_s$}-subproblem}: With the other variables fixed, the optimization problem associated with {\small$\mathbf{H}$} is formulated as
\begin{eqnarray}
	\begin{split}
	\small
	&\min_{\mathbf{H}_s\geq{0}}	\|\mathbf{X}_s-\mathbf{D}_s^{(1)}\dots\mathbf{D}_s^{(l)}\mathbf{H}_s^{(l)}\|_F^2+\gamma\sum_{l=1}^{L}tr(\mathbf{H}_s^{(l)}\mathbf{L}_s\mathbf{H}_s^{(l)^\top})\\
	&~~~~~~~~~~~+\sum_{l=1}^{L}\Phi\left(\bm{\Lambda}_1^{(l)},[\mathbf{H}_s^{(l)},\mathbf{H}_t^{(l)}]-\mathbf{H}_s^{(l)}\mathbf{Z}^{(l)}-\mathbf{E}^{(l)}\right).
	\end{split}
	\label{eq2-01}
\end{eqnarray}

It can be observed that the updates of {\small$\mathbf{H}_s^{(l)} (l=1,\dots,L)$} in each level are independent. Thus, {\small$\mathbf{H}_s^{(l)}$} can be updated one by one. For {\small$\mathbf{H}_s^{(l)}$}, we optimize the following problem:
\begin{eqnarray}
\begin{split}
\small
&\min_{\mathbf{H}_s^{(l)}\geq{0}}	\|\mathbf{X}_s-{\rm\Theta}_s^{(l)}\mathbf{H}_s^{(l)}\|_F^2+\gamma tr(\mathbf{H}_s^{(l)}\mathbf{L}_s\mathbf{H}_s^{(l)^\top})\\
&~~~~~~~~~~~+\frac{\mu}{2}\|[\mathbf{H}_s^{(l)},\mathbf{H}_t^{(l)}]-\mathbf{H}_s^{(l)}\mathbf{Z}^{(l)}-\mathbf{E}^{(l)}+{\bm{\Lambda}_1^{(l)}}/{\mu}\|_F^2.
\end{split}
\label{eq2-02}
\end{eqnarray}

Denoting {\small$\mathbf{E}^{(l)}=[\mathbf{E}_s^{(l)},\mathbf{E}_t^{(l)}]$}, {\small$\mathbf{Z}^{(l)}=[\mathbf{Z}_s^{(l)},\mathbf{Z}_t^{(l)}]$}, and {\small$\bm{\Lambda}_{1}^{(l)}=[\bm{\Lambda}_{1,s}^{(l)},\bm{\Lambda}_{1,t}^{(l)}]$}, it is equivalent to optimizing
\begin{eqnarray}
\begin{split}
\small
&\min_{\mathbf{H}_s^{(l)}\geq{0}}	\|\mathbf{X}_s-{\rm\Theta}_s^{(l)}\mathbf{H}_s^{(l)}\|_F^2+\gamma tr(\mathbf{H}_s^{(l)}\mathbf{L}_s\mathbf{H}_s^{(l)^\top})\\
&~~~~~~~~~~~+\frac{\mu}{2}\|\mathbf{H}_s^{(l)}-\mathbf{H}_s^{(l)}\mathbf{Z}_s^{(l)}-\mathbf{E}_s^{(l)}+{\bm{\Lambda}_{1,s}^{(l)}}/{\mu}\|_F^2.
\end{split}
\label{eq2-021}
\end{eqnarray}

By taking the derivative of Eq.~(\ref{eq2-021}) \emph{w.r.t.} {\small$\mathbf{H}_s^{(l)}$} and using the KKT condition \cite{boyd2004convex}, we have the updating rule:
\begin{equation}
\small
%\footnotesize
\begin{split}
&\mathbf{H}_s^{(l)}\leftarrow\mathbf{H}_s^{(l)}\odot\\
&\frac{2{{\rm\Theta}_s^{(l)}}^{\top}\mathbf{X}_s+\mu(\mathbf{E}_s^{(l)}-{\bm{\Lambda}_{1,s}^{(l)}}/{\mu})(\mathbf{I}-\mathbf{Z}_s^{(l)})^{\top}}{2{{\rm\Theta}_s^{(l)}}^{\top}{{\rm\Theta}_s^{(l)}}\mathbf{H}_s^{(l)}+\mu\mathbf{H}_s^{(l)}(\mathbf{I}-\mathbf{Z}_s^{(l)})(\mathbf{I}-\mathbf{Z}_s^{(l)})^{\top}+2\gamma\mathbf{H}_s\mathbf{L}_s},
\end{split}
\label{eq2-03}
\end{equation}
\noindent where {\small$\mathbf{I}$} denotes an identity matrix.

\textbf{{\small$\mathbf{H}_t$}-subproblem}: To update {\small$\mathbf{H}_t$}, the following objective function should be optimized:
\begin{eqnarray}
	\begin{split}
	\small
	&\min_{\mathbf{H}_t\geq{0}}	\|\mathbf{X}_t-\mathbf{D}_t^{(1)}\dots\mathbf{D}_t^{(l)}\mathbf{H}_t^{(l)}\|_F^2\\
	&~~~~~~~~~~~+\sum_{l=1}^L\Phi\left(\bm{\Lambda}_{1,t}^{(l)},\mathbf{H}_t^{(l)}-\mathbf{H}_s^{(l)}\mathbf{Z}_t^{(l)}-\mathbf{E}_t^{(l)}\right),
	\end{split}
	\label{eq2-041}
\end{eqnarray}

We also update {\small$\mathbf{H}_t^{(l)} (l=1,\dots,L)$} one by one. For {\small$\mathbf{H}_t^{(l)}$},
it is equivalent to optimizing the following problem:
\begin{eqnarray}
\begin{split}
\small
&\min_{\mathbf{H}_t^{(l)}\geq{0}}	\|\mathbf{X}_t-{\rm\Theta}_t^{(l)}\mathbf{H}_t^{(l)}\|_F^2\\
&~~~~~~~~~~~+\frac{\mu}{2}\|\mathbf{H}_t^{(l)}-\mathbf{H}_s^{(l)}\mathbf{Z}_t^{(l)}-\mathbf{E}_t^{(l)}+{\bm{\Lambda}_{1,t}^{(l)}}/{\mu}\|_F^2.
\end{split}
\label{eq2-04}
\end{eqnarray}

By taking the derivative of Eq.~(\ref{eq2-04}) \emph{w.r.t.} {\small$\mathbf{H}_t^{(l)}$} and using the KKT condition \cite{boyd2004convex}, the updating rule is given as:
\begin{equation}
\small
\begin{split}
\mathbf{H}_t^{(l)}&\leftarrow\mathbf{H}_t^{(l)}\odot\frac{2{{\rm\Theta}_t^{(l)}}^{\top}\mathbf{X}_t+\mu(\mathbf{H}_s^{(l)}\mathbf{Z}_t^{(l)}+\mathbf{E}_t^{(l)}-\frac{\bm{\Lambda}_{1,t}^{(l)}}{\mu})}{2{{\rm\Theta}_t^{(l)}}^{\top}{{\rm\Theta}_t^{(l)}}\mathbf{H}_t^{(l)}+\mu\mathbf{H}_t^{(l)}}.
\end{split}
\label{eq2-06}
\end{equation}

\textbf{{\small$\mathbf{J}$}-subproblem}: Dropping the unrelated terms, the optimization for  {\small$\mathbf{Z}^{(l)}$} yields
\begin{eqnarray}
\small
\begin{split}
\min_{\mathbf{J}^{(l)}}~&\Phi\left(\bm{\Lambda}_1^{(l)},[\mathbf{H}_s^{(l)},\mathbf{H}_t^{(l)}]-\mathbf{H}_s^{(l)}\mathbf{J}^{(l)}-\mathbf{E}^{(l)}\right)\\
&~~~~~~+\Phi\left(\bm{\Lambda}_2^{(l)},\mathbf{Z}^{(l)}-\mathbf{J}^{(l)}\right).
\end{split}
\label{eq7-01}
\end{eqnarray}

This is equivalent to optimizing the following problem:
\begin{eqnarray}
\small
\begin{split}
\min_{\mathbf{J}^{(l)}}~&\|[\mathbf{H}_s^{(l)},\mathbf{H}_t^{(l)}]-\frac{\bm{\Lambda}_1^{(l)}}{\mu}-\mathbf{E}^{(l)}+\mathbf{H}_s^{(l)}\mathbf{J}^{(l)}\|_F^2\\
&+\|\mathbf{J}^{(l)}-\mathbf{Z}^{(l)}-{\bm{\Lambda}_2^{(l)}}/{\mu}\|_F^2.
\end{split}
\label{eq7-02}
\end{eqnarray}

By taking the derivative of (\ref{eq7-02}) {\small$\emph{w.r.t}$} {\small$\mathbf{J}^{(l)}$} and setting it to zero, its closed-form solution is given as follows:
\begin{eqnarray}
\small
\begin{split}
\mathbf{J}^{(l)}=&\left(\mathbf{H}_s^{(l)^{\top}}\mathbf{H}_s^{(l)}+\mathbf{I}\right)^{-1}\\
&\mathbf{H}_s^{(l)^{\top}}\left([\mathbf{H}_s^{(l)},\mathbf{H}_t^{(l)}]+\frac{\bm{\Lambda}_1^{(l)}}{\mu}-\mathbf{E}^{(l)}\right)+\left(\mathbf{Z}^{(l)}+\frac{\bm{\Lambda}_2^{(l)}}{\mu}\right).
\end{split}
\label{eq7-03}
\end{eqnarray}

\textbf{{\small$\mathbf{Z}$}-subproblem}: By dropping the unrelated terms, we can optimize the following problem for updating {\small$\mathbf{Z}^{(l)}$}:
\begin{eqnarray}
\small
\begin{split}
\min_{\mathbf{Z}^{(l)}}~&\beta\sum_{m\neq{l}}^L\textup{HSIC}\left(\mathbf{Z}^{(l)},\mathbf{Z}^{(m)}\right)+\gamma\left(tr(\mathbf{J}^{(l)}\mathbf{L}\mathbf{Z}^{(l)^{\top}}\right)\\
&+\Phi(\bm{\Lambda}_2^{(l)},\mathbf{Z}^{(l)}-\mathbf{J}^{(l)}),~~s.t.~\mathbf{Z}^{(l)}\geq{0},\mathbf{1}^{\top}\mathbf{Z}^{(l)}=\mathbf{1}^{\top}.
\end{split}
\label{eq5-01}
\end{eqnarray}

In this study, we adopt the inner product kernel for the HSIC constraint, \ie, {\small$\mathbf{K}_l={\mathbf{Z}^{(l)}}^{\top}\mathbf{Z}^{(l)}$}. Then, we have {\small$\textup{HSIC}(\mathbf{Z}^{(l)},\mathbf{Z}^{(m)})=tr(\mathbf{Z}^{(l)}\mathbf{K}{\mathbf{Z}^{(l)}}^{\top})$} with {\small$\mathbf{K}=\sum_{m\neq{l}}^L\mathbf{M}\mathbf{K}_m\mathbf{M}$}, where {\small$m_{ij}=\delta_{ij}-1/n$}. Finally, by taking the derivative of (\ref{eq5-01}) {\small$\emph{w.r.t}$} {\small$\mathbf{Z}^{(l)}$}, we can obtain
\begin{eqnarray}
\begin{split}
&\frac{\partial	\mathcal{J}(\mathbf{Z}^{(l)})}{\partial	{\mathbf{Z}^{(l)}}}=\beta\mathbf{Z}^{(l)}\mathbf{K}+\gamma\mathbf{Z}^{(l)}\mathbf{L}+\frac{\mu}{2}\mathbf{H}_s^{(l)^{\top}}\mathbf{H}_s^{(l)}\mathbf{Z}^{(l)}\\
&~~~~~~~~~~~~~~~~~~-\mathbf{H}_s^{(l)^{\top}}\left([\mathbf{H}_s^{(l)},\mathbf{H}_t^{(l)}]+\frac{\bm{\Lambda}_1^{(l)}}{\mu}-\mathbf{E}^{(l)}\right).
\end{split}
\label{eq5-03}
\end{eqnarray}

By setting {\small$\frac{\partial	\mathcal{J}(\mathbf{Z}^{(l)})}{\partial	{\mathbf{Z}^{(l)}}}$} to zero, we have its closed-form solution. After that, an iterative algorithm \cite{huang2015new} is used to obtain the optimal solution of {\small$\mathbf{Z}^{(l)}$}.

\textbf{{\small$\mathbf{E}$}-subproblem}: Updating the error term {\small$\mathbf{E}^{(l)}$} is equivalent to solving the following problem:
\begin{equation}
\small
\begin{split}
\min_{\mathbf{E}^{(l)}}~\frac{1}{\mu}\|\mathbf{E}^{(l)}\|_{2,1}+\frac{1}{2}\|\mathbf{E}^{(l)}-\mathbf{G}\|_F^2,
\end{split}
\label{eq6-01}
\end{equation}
\noindent where {\small$\mathbf{G}=[\mathbf{H}_s^{(l)},\mathbf{H}_t^{(l)}]-\mathbf{H}_s^{(l)}\mathbf{Z}^{(l)}+\bm{\Lambda}_1^{(l)}/\mu$}. This optimization problem is solved using the algorithm in \cite{liu2012robust}.

\textbf{Multipliers updating}: The multipliers {\small$\bm{\Lambda}_1^{(l)}$} and {\small$\bm{\Lambda}_2^{(l)}$} ($l=1,2,\dots,L$) can be updated by
\begin{equation}
\small
\left\{
\begin{aligned}
&\bm{\Lambda}_1^{(l)}:=\bm{\Lambda}_1^{(l)} + \mu([\mathbf{H}_s^{(l)},\mathbf{H}_t^{(l)}]-\mathbf{H}_s^{(l)}\mathbf{Z}^{(l)}-\mathbf{E}^{(l)}),\\
&\bm{\Lambda}_2^{(l)}:=\bm{\Lambda}_2^{(l)} + \mu(\mathbf{Z}^{(l)}-\mathbf{J}^{(l)}).\\
\end{aligned}
\right.
\label{eq91}
\end{equation}

\noindent\textbf{Initialization and implementation}: Following previous deep NMF-based works~\cite{trigeorgis2016deep,zhao2017multi}, we first pretrain a deep NMF model to obtain initial approximations for {\small$\mathbf{D}_s^{(l)}$}, {\small$\mathbf{D}_t^{(l)}$}, {\small$\mathbf{H}_s^{(l)}$}, and {\small$\mathbf{H}_t^{(l)}$} ($l=1,2,\dots,L$) in each layer. This pretraining process often reduces the training time of our model, and its effectiveness has also been proven in deep auto-encoder networks \cite{hinton2006reducing}. Let us consider the source data as an example. We decompose {\small$\mathbf{X}_s\approx\mathbf{D}_s^{(1)}\mathbf{H}_s^{(1)}$} and further decompose {\small$\mathbf{H}_s^{(1)}\approx\mathbf{D}_s^{(2)}\mathbf{H}_s^{(2)}$}, until all layers are initialized. Then, we carry out the optimization steps for all variables and repeat this until convergence. The detailed steps for optimizing the proposed framework in Eq.~(\ref{eq00-01}) via the ADMM algorithm are summarized in Algorithm~\ref{alg01}.

\renewcommand\arraystretch{0.9}
\begin{algorithm}[!t]
	\footnotesize
	\caption{\footnotesize Optimizing the problem (\ref{eq00-01}) via ADMM.}	\label{alg01}
	\SetKwInOut{Input}{Input}\SetKwInOut{Output}{Output}
	\textbf{Input:} {Source data: $\mathbf{X}_{s}$ and target data $\mathbf{X}_{t}$, parameters $\alpha$, $\lambda$, $\beta$, and $\gamma$.}\\
	\textbf{Initialize:} $\bm{\Lambda}_1^{(l)}=0$, $\bm{\Lambda}_2^{(l)}=0$, $\varepsilon=10^{-4}$, $\rho=1.5$, $\mu=10^{-4}$, $max_{\mu}=10^{6}$.\\
	\textbf{Output:}{ ${\textbf{Z}^{(l)}}, l=1,2,\dots,L$.}\\
	\While{not converged}{
		\For{l=1,2,\ldots,L}{
			
			Update $\mathbf{D}_s^{(l)}$, $\mathbf{D}_t^{(l)}$, $\mathbf{H}_s^{(l)}$, $\mathbf{H}_t^{(l)}$, $\mathbf{J}^{(l)}$, $\mathbf{Z}^{(l)}$, $\mathbf{E}^{(l)}$, $\bm{\Lambda}_1^{(l)}$, and $\bm{\Lambda}_2^{(l)}$ using Eqs.~(\ref{eq1-032}), (\ref{eq1-033}), (\ref{eq2-03}), (\ref{eq2-06}), (\ref{eq7-03}), (\ref{eq6-01}), and (\ref{eq91}), respectively.\\

		}
		
		Update the parameter $\mu$ via $\mu=\min(\rho\mu,max_{\mu})$;\\
		Check the convergence conditions:\\
		~~~$\|[\mathbf{H}_s^{(l)},\mathbf{H}_t^{(l)}]-\mathbf{H}_s^{(l)}\mathbf{Z}^{(l)}-\mathbf{E}^{(l)}\|_{\infty}<\varepsilon$ \\
		~~~and $\|\mathbf{Z}^{(l)}-\mathbf{J}^{(l)}\|_{\infty}<\varepsilon$.
	}
\end{algorithm}

\subsection{Clustering-based Motion Segmentation}
\label{clustering}

With Algorithm~\ref{alg01}, we can obtain the learned multi-level representations~{\small$\mathbf{Z}^{(l)} (l=1,2,\dots,L)$}. Then, we extract the corresponding target representations {\small$\mathbf{Z}_t^{(l)}\in\mathbb{R}^{n_s\times{n_t}}$} from {\small$\mathbf{Z}^{(l)}=[\mathbf{Z}_s^{(l)},\mathbf{Z}_t^{(l)}]$}.
Inspired by \cite{li2015temporal}, we develop a new similarity measurement to construct an affinity matrix {\small$\mathbf{A}$} for our multi-level transfer subspace learning, which can explore intrinsic relationships among within-cluster samples from human motion data.
Specifically, each element of {\small$\mathbf{A}$} can be defined as the distance between a pair of the learned representation coefficients for target data, which is given by
\begin{equation}
\small
\begin{aligned}
a_{ij}=\frac{1}{L}\sum\nolimits_{l=1}^L\frac{{\mathbf{z}_{t,i}^{(l)}}^{\top}\mathbf{z}_{t,j}^{(l)}}{\|\mathbf{z}_{t,i}^{(l)}\|_2\|\mathbf{z}_{t,j}^{(l)}\|_2},
\end{aligned}
\label{eq-32}
\end{equation}
\noindent where $\mathbf{z}_{t,i}^{(l)}$ and $\mathbf{z}_{t,j}^{(l)}$ denote the $i$-th and $j$-th columns of $\mathbf{z}_{t}^{(l)}$, respectively. Note that we compute each element $a_{ij}$ by averaging the pair distance among multi-level feature spaces. After that, the Normalized Cuts \cite{shi1998motion} algorithm is applied to the learned affinity matrix $\mathbf{A}$ to produce the clustering results.

\subsection{Complexity Analysis}
\label{complexity}

The overall optimization procedure consists of two stages, \ie, pretraining and model updating, thus we analyze the computational burden of the two parts separately. For clarity, we define {\small$p$} as the maximal layer size in all layers and $n$ ($n=n_s+n_t$) as the total number of the source and target data. For the pretraining part, the computational complexity is of order {\small$\mathcal{O}(Lt_p(n_s^2p+n_sp^2+n_t^2p+n_tp^2))$}, where {\small$t_p$} is the number of iterations. For the model updating part, the computational cost lies in updating {\small$\mathbf{D}_s^{(l)}$}, {\small$\mathbf{D}_t^{(l)}$}, {\small$\mathbf{H}_s^{(l)}$, $\mathbf{H}_t^{(l)}$}, {\small$\mathbf{J}^{(l)}$}, {\small$\mathbf{Z}^{(l)}$}, and {\small$\mathbf{E}^{(l)}$}. Thus the computational complexity is of order {\small$\mathcal{O}(Lt_u(n_s^2p+n_sp^2+n_t^2p+n_tp^2+p^3+n^3+n_s^2{n}+n_{s}n_{t}p))$}, where {\small$t_u$} is the number of iterations in this stage. Finally, considering {\small$n_s,n_t>{p}$} in our task, the overall computational cost is {\small$\mathcal{O}(L((t_p+t_u)(n_s^2p+n_sp^2+n_t^2p+n_tp^2)+t_u(n^3+n_{s}n_{t}p)))$}.

\begin{table*}[!htp]
	\caption{\footnotesize Clustering results of compared methods in terms of NMI and ACC on four human motion datasets. Names in brackets indicate the source datasets. M, K, W, and U denote MAD, Keck, Weizmann, and UT-interaction, respectively. The first two best results are highlighted in \textbf{bold} and \underline{underlined} when using the same source data.}
	\scriptsize
		\begin{minipage}{0.25\linewidth}
			\centering
			\renewcommand\arraystretch{1.3}
			{
				(a) Results on Keck dataset\vspace{-0.15cm}
				\begin{tabular}[t]{l|c|c}
					\hline
					\rowcolor{mygray} Method & NMI $\uparrow$ & ACC $\uparrow$\\
					\hline
					SC \cite{ng2002spectral}         & 0.4744    & 0.3886   \\
					KMD \cite{rdusseeun1987clustering}       &0.4702    &0.3970\\
					LRR \cite{liu2013robust}         & 0.4862   & 0.4297   \\
					OSC \cite{tierney2014subspace}        & 0.5931    & 0.4393   \\
					SSC \cite{elhamifar2013sparse}         & 0.3858    & 0.3137   \\
					LSR \cite{lu2012robust}          & 0.4548    & 0.4894  \\
					\hline

					TSC(M)  \cite{li2015temporal}   & 0.6935    & 0.4653  \\
					TSS(M)  \cite{wang2018learning}    & 0.8049    & 0.5395   \\
					LTS(M) \cite{wang2018low}    & \underline{0.8226}     & 0.5509  \\

					MCSTL(M)  \cite{zhoucvpr}   & \textbf{0.8270}  & \underline{0.6010} \\
					CDMS(M)    & {0.7891}  & \textbf{0.6044} \\
					\hline
					TSC(W) \cite{li2015temporal}    & 0.6862    & 0.4548  \\
					TSS(W) \cite{wang2018learning}   & 0.7928     & 0.5485  \\
					LTS(W) \cite{wang2018low}      & 0.7983     & 0.5649  \\

					MCSTL(W) \cite{zhoucvpr}   & \underline{0.8196}  & \underline{0.5915} \\
					CDMS(W)   & \textbf{0.8213}  & \textbf{0.6085} \\
					\hline	
					TSC (U) \cite{li2015temporal}    & 0.6797    & 0.4421  \\
					TSS(U)  \cite{wang2018learning}   & 0.7937     & 0.4951 \\
					LTS(U)  \cite{wang2018low}     & 0.7947      & 0.5519  \\

					MCSTL(U) \cite{zhoucvpr}    & \textbf{0.8120}  & \underline{0.6105} \\
					CDMS(U)    & \underline{0.8040}  & \textbf{0.6207} \\
					\hline
			\end{tabular}}
		\end{minipage}
		\begin{minipage}{0.25\linewidth}
			\centering
			\renewcommand\arraystretch{1.3}

			{
				(b) Results on MAD dataset\vspace{-0.15cm}
				\begin{tabular}[t]{l|c|c}

					\hline

					\rowcolor{mygray} Method & NMI $\uparrow$  & ACC $\uparrow$\\
					\hline
					SC   \cite{ng2002spectral}          & 0.4369     & 0.3639   \\
					KMD  \cite{rdusseeun1987clustering} &0.3914      &0.3226\\
					LRR  \cite{liu2013robust}           & 0.2249     & 0.2397    \\
					OSC  \cite{tierney2014subspace}     & 0.5589     & 0.4327  \\
					SSC  \cite{elhamifar2013sparse}     & 0.4758     & 0.3817  \\
					LSR  \cite{lu2012robust}            & 0.3667     & 0.3979 \\
					
					\hline
					TSC(K)  \cite{li2015temporal}     & 0.7691    & 0.5473 \\
					TSS(K)  \cite{wang2018learning}   & \textbf{0.8286}    & 0.5792   \\
					LTS(K)  \cite{wang2018low}      & 0.8244   & 0.5874  \\

					MCSTL(K) \cite{zhoucvpr}    & 0.8099 & \underline{0.6125} \\
					CDMS(K)    & \underline{0.8251} & \textbf{0.6536} \\
					
					\hline	
					TSC(W)  \cite{li2015temporal}  & 0.7684   & 0.5418  \\
					TSS(W)  \cite{wang2018learning}  & 0.8202   & 0.5736 \\
					LTS(W)  \cite{wang2018low}     & {0.8213}   & 0.5906 \\

					MCSTL(W) \cite{zhoucvpr}  & \textbf{0.8307} & \underline{0.6158} \\
					CDMS(W)   & {0.8238} & \textbf{0.6392} \\
					
					\hline	
					TSC (U) \cite{li2015temporal}   & 0.7691    & 0.5315  \\
					TSS(U)  \cite{wang2018learning}   & 0.8108    & 0.5479 \\
					LTS(U)  \cite{wang2018low}     & {0.8211}     & 0.5980 \\

					MCSTL(U) \cite{zhoucvpr}    & \textbf{0.8314} & \underline{0.6163}\\
					CDMS(U)   & \underline{0.8238} & \textbf{0.6371} \\
					\hline
			\end{tabular}}
		\end{minipage}
		\begin{minipage}{0.25\linewidth}
			\centering
			\renewcommand\arraystretch{1.3}

			{
				(c) Results on Weizman dataset\vspace{-0.15cm}
				\begin{tabular}[t]{l|c|c}

					\hline

					\rowcolor{mygray} Method & NMI $\uparrow$  & ACC $\uparrow$\\
					\hline
					SC   \cite{ng2002spectral}            & 0.5435    & 0.4127  \\
					KMD  \cite{rdusseeun1987clustering}   &0.5289    &0.4441\\
					LRR  \cite{liu2013robust}             & 0.4382    & 0.3638    \\
					OSC  \cite{tierney2014subspace}       & 0.7047     & 0.5216   \\
					SSC  \cite{elhamifar2013sparse}       & 0.6009   & 0.4576   \\
					LSR  \cite{lu2012robust}              & 0.5093    & 0.5091   \\
					
					\hline
					TSC(K)  \cite{li2015temporal}    & 0.7971     & 0.5931  \\
					TSS(K)  \cite{wang2018learning}  & 0.8326    & 0.6030   \\
					LTS(K)  \cite{wang2018low}       & \underline{0.8599}    & 0.6391  \\
					MCSTL(K) \cite{zhoucvpr}    & 0.8371   & \underline{0.6436} \\
					CDMS(K)    & \textbf{0.8601}   & \textbf{0.6465} \\
					
					\hline
					TSC(M)   \cite{li2015temporal}    & 0.8032    & 0.5961  \\
					TSS(M)   \cite{wang2018learning}  & \underline{0.8509}    & 0.6208  \\
					LTS(M)   \cite{wang2018low}       & \textbf{0.8579}    & 0.6156  \\
					MCSTL(M) \cite{zhoucvpr}   &  0.8232  & \underline{0.6348}\\
					CDMS(M)    & {0.8375}   & \textbf{0.6505} \\
					
					\hline
					TSC (U)  \cite{li2015temporal}    & 0.7796     & 0.5402  \\
					TSS(U)   \cite{wang2018learning}  & 0.8124     & 0.5865  \\
					LTS(U)   \cite{wang2018low}       & 0.8267    & 0.6122  \\
					MCSTL(U) \cite{zhoucvpr}   & \underline{0.8351}   & \textbf{0.6371} \\
					CDMS(U)    & \textbf{0.8616}   & \underline{0.6266} \\
					\hline
			\end{tabular}}
		\end{minipage}
		\begin{minipage}{0.2\linewidth}
			\centering
			\renewcommand\arraystretch{1.3}
			{
				(d) Results on UT dataset\vspace{-0.15cm}
				\begin{tabular}[t]{l|c|c}
					
					\hline
					\rowcolor{mygray} Method & NMI $\uparrow$  & ACC $\uparrow$\\
					\hline
					SC   \cite{ng2002spectral}            & 0.4894     & 0.4477  \\
					KMD  \cite{rdusseeun1987clustering}   & 0.5108     & 0.5122\\
					LRR  \cite{liu2013robust}             & 0.4051     & 0.4162    \\
					OSC  \cite{tierney2014subspace}       & 0.6877     & 0.5846  \\
					SSC  \cite{elhamifar2013sparse}       & 0.4998     & 0.4389   \\
					LSR   \cite{lu2012robust}             & 0.4322     & 0.5183  \\

					\hline
					TSC(K)  \cite{li2015temporal}      & 0.7216    & 0.5213 \\
					TSS(K)  \cite{wang2018learning}    & 0.7746    & 0.5371   \\
					LTS(K)  \cite{wang2018low}         & 0.7961    & 0.6127 \\
					MCSTL(K)  \cite{zhoucvpr}    & \underline{0.8121}  & \underline{0.6148} \\
					CDMS(K)      & \textbf{0.8267}     & \textbf{0.6547} \\
					
					\hline
					TSC(M)   \cite{li2015temporal}     & 0.7442    & 0.5288  \\
					TSS(M)   \cite{wang2018learning}   & 0.7783    & 0.5335  \\
					LTS(M)   \cite{wang2018low}        & 0.8128    & 0.6299  \\
					MCSTL(M)  \cite{zhoucvpr}   &\underline{ 0.8239}      & \underline{0.6433} \\
					CDMS(M)     & \textbf{0.8306}   & \textbf{0.6643} \\
					
					\hline
					TSC (W)  \cite{li2015temporal}   & 0.7136    & 0.5111  \\
					TSS(W)   \cite{wang2018learning} & 0.7878    & 0.5944 \\
					LTS(W)   \cite{wang2018low}      & 0.8035    & 0.6296 \\
					MCSTL(W) \cite{zhoucvpr}    & \underline{0.8198}       & \underline{0.6463} \\
					CDMS(W)     & \textbf{0.8288}    & \textbf{0.6642} \\
					\hline
			\end{tabular}}
		\end{minipage}
	
	\label{tab02}
\end{table*}

\section{Experimental Results and Analysis}
\label{Experiments}

In this section, we first introduce the human motion datasets used in our experiments (\secref{dataset}), and provide the details of the experimental setup (\secref{setup}). Then we show the comparison results with several state-of-the-art methods (\secref{comparison}) and conduct the model study (\secref{study}).

\subsection{Human Motion Datasets}
\label{dataset}

\begin{figure}[t]
	\begin{center}
		\includegraphics[width=0.5\textwidth]{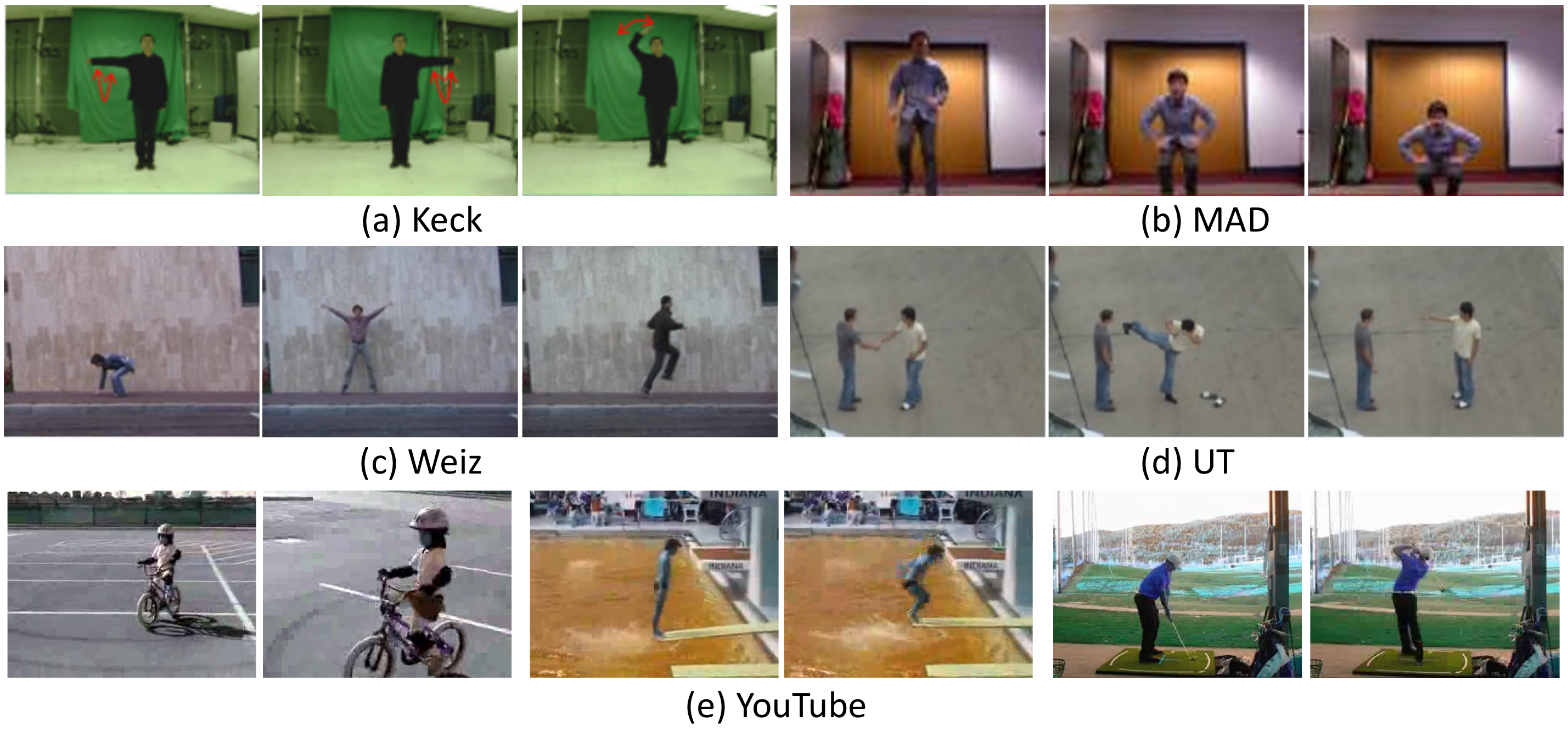}\vspace{-0.25cm}
		\caption{\footnotesize Sampling frames from five human motion benchmark datasets, \ie, (a) Keck, (b) MAD, (c) Weiz, (d) UT, and (e) YouTube.}\vspace{-0.45cm}
		\label{fig04}
	\end{center}
\end{figure}

To comprehensively evaluate the proposed model, we conduct experiments on five benchmark human motion datasets (see Fig.~\ref{fig04} for some example frames). $\bullet$ \textbf{Keck Gesture Dataset (Keck)} \cite{jiang2012recognizing} consists of 14 different actions from military signals, in which each subject is carried out 14 actions and gestures. Besides, the videos in this dataset were obtained by a fixed camera when these subjects stand out in a static background. $\bullet$ \textbf{Multi-Modal Action Detection Dataset (MAD)} \cite{huang2014sequential} consists of actions captured from various modalities using a Microsoft Kinect V2 system, which includes RGB images, depth cues, and skeleton formats. Specifically, the RGB images and 3D depth cues are of a size of $240\times320$. Moreover, each subject performs 35 different actions within two indoor scenes. $\bullet$ \textbf{Weizmann Dataset (Weiz)} \cite{gorelick2007actions} is composed of 90 video sequences with 10 actions (running, walking, skipping, bending, etc.) captured by nine subjects in an outdoor environment. All videos have a size of $180\times144$ with 50 fps. $\bullet$ \textbf{UT-Interaction Dataset (UT)} \cite{ryoo2009spatio} is composed of 20 videos, each of which includes six different action types of human-human interactions (such as punching, pushing, pointing, hugging, kicking, and handshaking). $\bullet$ \textbf{Youtube Dataset} \cite{liu2012learning} consists of 11 action categories, which are diving, biking/cycling, horse back riding, golf swinging, soccer juggling, swinging, walking, trampoline jumping, volleyball spiking, tennis swinging, and basketball shooting. In our experiment, we construct four sequences, each of which includes the first 10 actions.

\subsection{Experimental Setup}
\label{setup}

\subsubsection{Dataset Settings}

Following the dataset preprocessing and settings in \cite{wang2018low}, in our experiments, we use the extracted $324$-dimensional HOG features \cite{danal2005histgram} for each frame. Due to the fact that different datasets consist of different numbers of actions, we randomly choose ten motions from the Keck, MAD, and Weiz datasets in our experiments. In addition, we extract deep CNN features from the Keck, Weiz, and Youtube datasets using the pretrained VGG-16 \cite{simonyan2014very} model and obtain a $1000$-dimension feature vector for each frame. For model evaluation, we select one dataset as the source and the remaining datasets as the target, and we obtain the mean results of motion segmentation on the target dataset.

\subsubsection{Compared Methods}

We compare the proposed model with several state-of-the-art approaches, which include (1) Spectral clustering (\textbf{SC}) \cite{ng2002spectral}. The feature vectors of target samples are directly fed into the standard spectral clustering algorithm \cite{ng2002spectral} to obtain clustering results. (2) K-medoids (\textbf{KMD}) \cite{rdusseeun1987clustering} selects target samples as centers and partitions them into different groups using the Manhattan Norm to measure the distance between data points. (3) Low-rank representation (\textbf{LRR}) \cite{liu2013robust} conducts subspace clustering by imposing a low-rank constraint on the target data. (4) Ordered subspace clustering (\textbf{OSC}) \cite{tierney2014subspace} utilizes a temporal correlation constraint to force representations of the temporal data to be closer. (5) Sparse subspace clustering (\textbf{SSC}) \cite{elhamifar2013sparse} assumes that there exists a dictionary that can be used to linearly represent all data points, and a sparse constraint is imposed on the representation coefficients. (6) Least square regression (\textbf{LSR}) \cite{lu2012robust} introduces the $\ell_{2}$ regularization on the learned coefficient matrix, in which the grouping effect of LSR enables to group mostly correlated data together. (7) Temporal subspace clustering (\textbf{TSC}) \cite{li2015temporal} employs a non-negative dictionary learning algorithm to obtain an expressive coefficient matrix for temporal clustering, and a temporal Laplacian regularization is imposed on the coefficient matrix. (8) Transfer subspace segmentation (\textbf{TSS}) \cite{wang2018learning} is a transferable subspace clustering method that explores useful information from relevant source data to boost clustering performance on target human motion data. (9) Low-rank transfer subspace (\textbf{LTS}) \cite{wang2018low} employs a graph regularizer to capture temporal correlation in both the source and target data as well as uncovers clustering structures within data by introducing a weighted low-rank constraint. For convenience, we denote the preliminary of our method, proposed in ~\cite{zhoucvpr}, as MCSTL.

\subsubsection{Evaluation Metrics and Parameter Settings}

To evaluate the model performance, we utilize two popular metrics, including \emph{i.e.}, normalized mutual information (NMI) and accuracy (ACC). Specifically, NMI evaluates the mutual information between the ground truth and the recovered cluster labels, and ACC computes the classification score with the best map. Note that, higher values for both metrics indicate better performance.
In addition, for our approach, we tune three trade-off parameters $\alpha$, $\beta$, and $\gamma$ in the range of $\{10^{-5},10^{-4},\ldots,10^{2}\}$. Furthermore, the number of layers for our CDMS model is set as $3$, and the number of sequential neighbors is tuned from the set of $\{9,11,\dots,21\}$.

\begin{table*}[!htp]
	\caption{\footnotesize Clustering results of compared methods in terms of NMI and ACC on three human motion datasets. Names in brackets denote the source datasets, where K, W, and Y denote Keck, Weizmann, and Youtube datasets, respectively. The first two best clustering results are highlighted in \textbf{bold} and \underline{underlined} when using the same source data. }
	\scriptsize
		\begin{minipage}{0.32\linewidth}
			\centering
			\renewcommand\arraystretch{1.3}
			{
				(a) Results on Keck dataset\vspace{-0.15cm}
				\begin{tabular}[t]{l|c|c}
					\hline
					\rowcolor{mygray} Method & NMI $\uparrow$ & ACC $\uparrow$\\
					\hline

					\hline
					TSS(W)   \cite{wang2018learning}   & 0.8595    & 0.5675   \\
					LTS(W)   \cite{wang2018low}        & 0.7906    & 0.5730  \\
					MCSTL(W)  \cite{zhoucvpr}   & \textbf{0.8675}      & \textbf{0.6423} \\
					CDMS(W)     & \underline{0.8663}   & \underline{0.5922} \\
					
					\hline
					TSS(Y)   \cite{wang2018learning}  &0.8050     & 0.5471      \\
					LTS(Y)   \cite{wang2018low}       & 0.8167    & 0.5690  \\
					MCSTL(Y) \cite{zhoucvpr}   & \underline{0.8235}  & \underline{0.5875}\\
					CDMS(Y)    & \textbf{0.8333}   & \textbf{0.5940} \\
					\hline

			\end{tabular}}
		\end{minipage}
		\begin{minipage}{0.32\linewidth}
			\centering
			\renewcommand\arraystretch{1.3}
			{
				(b) Results on Weiz dataset\vspace{-0.15cm}
				\begin{tabular}[t]{l|c|c}
					\hline

					\rowcolor{mygray} Method & NMI $\uparrow$  & ACC $\uparrow$\\
					\hline

					\hline

					TSS(K)  \cite{wang2018learning}    & 0.8189    & 0.5913   \\
					LTS(K)  \cite{wang2018low}         & \underline{0.8477}   & 0.6346  \\

					MCSTL(K) \cite{zhoucvpr}    & 0.7803 & \underline{0.6175} \\
					CDMS(K)    & \textbf{0.8539} & \textbf{0.6574} \\
					
					\hline	
					TSS(Y)   \cite{wang2018learning}  & \underline{0.8194}   & 0.5971 \\
					LTS(Y)   \cite{wang2018low}       & 0.8018   & 0.5906 \\

					MCSTL(Y)  \cite{zhoucvpr}  & 0.8111 & \underline{0.6066} \\
					CDMS(Y)    & \textbf{0.8579} & \textbf{0.6373}\\
					
					\hline	

			\end{tabular}}
		\end{minipage}
		\begin{minipage}{0.32\linewidth}
			\centering
			\renewcommand\arraystretch{1.3}

			{
				(c) Results on Youtube dataset\vspace{-0.15cm}
				\begin{tabular}[t]{l|c|c}
					\hline
					\rowcolor{mygray} Method & NMI $\uparrow$  & ACC $\uparrow$\\
					\hline

					\hline
					TSS(K)  \cite{wang2018learning}    & 0.8444   &0.6064 \\
					LTS(K)  \cite{wang2018low}         & \underline{0.8898}   &0.6226  \\
					MCSTL(K) \cite{zhoucvpr}    & 0.8141   & \underline{0.6440} \\
					CDMS(K)     & \textbf{0.9133}   & \textbf{0.6798} \\
					
					\hline
					TSS(W)  \cite{wang2018learning}   & 0.8820    & 0.6294  \\
					LTS(W)  \cite{wang2018low}        & 0.8695    & 0.6169  \\
					MCSTL(W) \cite{zhoucvpr}   & \underline{0.8839}   & \underline{0.6314}\\
					CDMS(W)    & \textbf{0.8863}   & \textbf{0.6420} \\
					
					\hline

			\end{tabular}}
		\end{minipage}
		
	\label{tab03}
\end{table*}

\subsection{Performance Comparison}
\label{comparison}

In all comparisons, we set one sequence as the source and another one as the target. As we use four datasets for our evaluations, we report the segmentation results when testing on one dataset at one time, using the remaining three as the source domains. Besides, since SC, KMD, LRR, OSC, SSC, and LSR are not designed to utilize source information, we only employ target videos as input for these methods. For the TSC, TSS, and LRT models, both source and target videos are fed for segmentation. The clustering results of different methods using HOG features are shown in Table~\ref{tab02}, where \textbf{bold} and \underline{underlined} denote the best two results.

From the results in Table~\ref{tab02}, we make the following observations:
(1) Compared with SC, KMD, LRR, OSC, SSC, and LRR, our method transfers more useful information from the source data to learn distinctive representations of the target data, resulting in improved segmentation performance. (2) Compared with transfer clustering-based segmentation methods (including TSC, TSS, and TSS), our method also obtains much better performance. This is because our approach explores domain-invariant features and preserves domain-specific knowledge simultaneously. These two aspects are equally important for transfer learning. More importantly, our model fuses multi-level representations to construct the affinity matrix for motion segmentation, which effectively exploits hierarchical semantics and structural information in a layer-wise fashion. Thus, these results validate the effectiveness of the proposed method against other state-of-the-art models in human motion segmentation.

\begin{figure}[t]
	\begin{center}
		\includegraphics[width=0.49\textwidth]{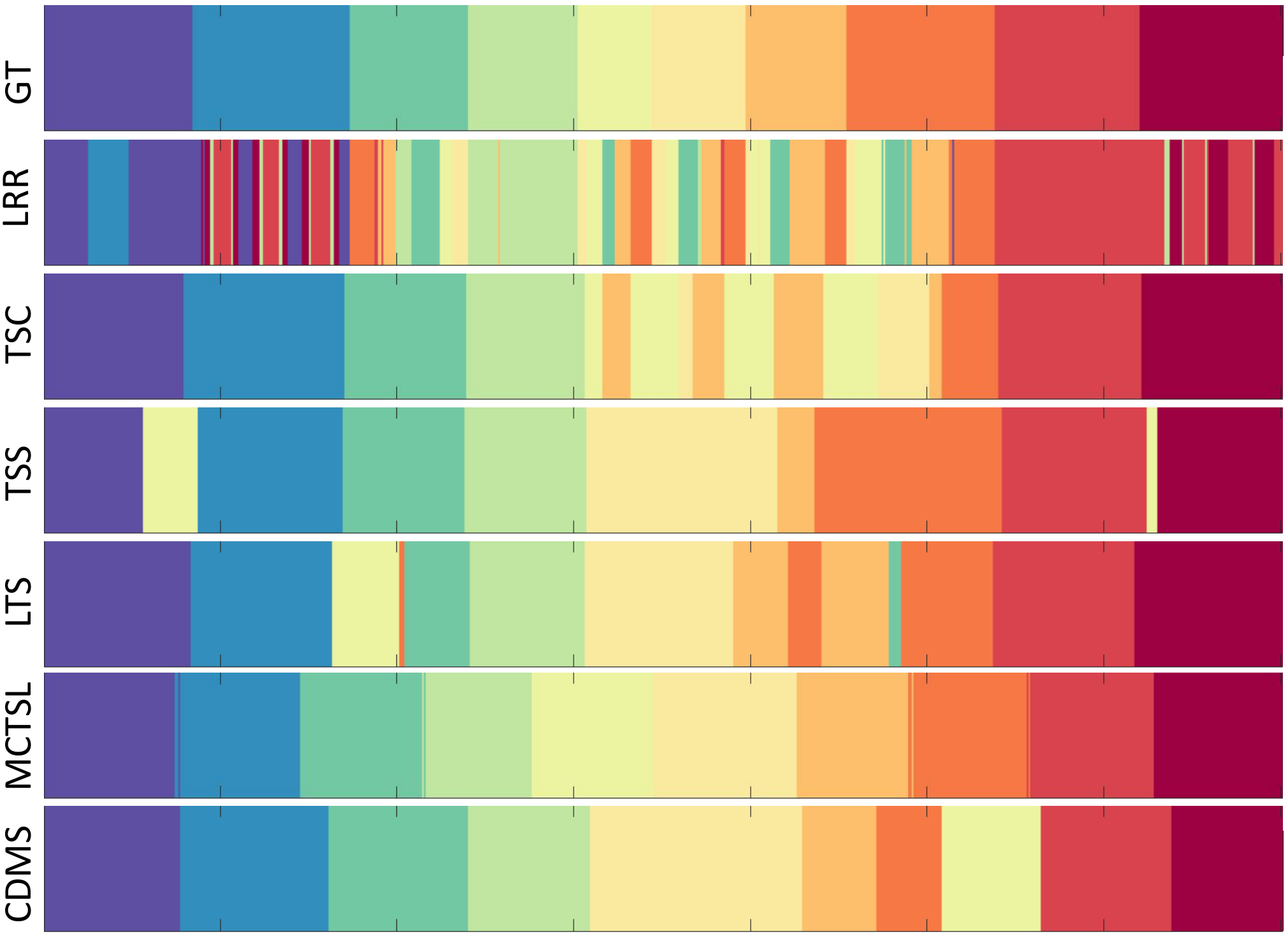}\vspace{-0.15cm}
		\caption{\footnotesize Visualization of clustering results on a sample video of the Weiz dataset. The ten colors denote ten different temporal clusters.}\vspace{-0.25cm}
		\label{fig05}
	\end{center}\vspace{-0.25cm}
\end{figure}

In Fig.~\ref{fig05}, we visualize the clustering results obtained by our model and other state-of-the-art methods on a sample video of the Weiz dataset. Different colors denote different clusters in the action data. From the results, it can be observed that the LRR and SSC models generate multiple fragments and cannot achieve meaningful temporal segmentation. The main reason is likely that they do not consider the temporal correlations when conducting subspace clustering. Compared with the LRR and SSC methods, TSC obtains a relatively better segmentation but it still suffers from many unexpected fragments. LTS and TSS obtain much better performance in most cases; however, they occasionally generate some unexpected fragments. Overall, our methods (CDMS and MCTSL) achieve continuous and meaningful segments, providing much better segmentation results than other comparison methods.

\textbf{Performance on Deep Features}. As deep models have demonstrated promising performance in several computer vision tasks, we further evaluate the performance of the proposed model using deep features as inputs. In this study, deep features are extracted using the pretrained VGG-16 \cite{simonyan2014very} model for each frame in the source and target datasets, and then we test the performance using the same settings as the previous experiments. We compare the proposed models (MCSTL and CDMS) with two state-of-the-art subspace transfer methods (\ie, TSS and LTS), and the comparison results are shown in Table~\ref{tab03}. From this experiment, we have the following observations: (1) our MSTL model performs better than other comparison methods in most cases; (2) the models achieve similar or slightly better performance using deep features rather than HOG features. Moreover, it can be noted that there is a difference in computation cost between using the two types of features. When using deep features, it needs a large-scale dataset to train a pre-trained model, and then we can extract deep features as the inputs. The dimension of deep features is often high (\eg, $1000$ in this study), which requires many computation resources. While using hand-crafted (\eg, HOG) features, it does not need extra training and requires fewer computation resources to process low-dimension features. Regardless of the input, our model is a general approach that can take hand-crafted features extracted from raw data or deep features.

\subsection{Model Study}
\label{study}

\subsubsection{Parameter Sensitivity}

\begin{figure}
	\begin{center}
		\includegraphics[width=0.5\textwidth]{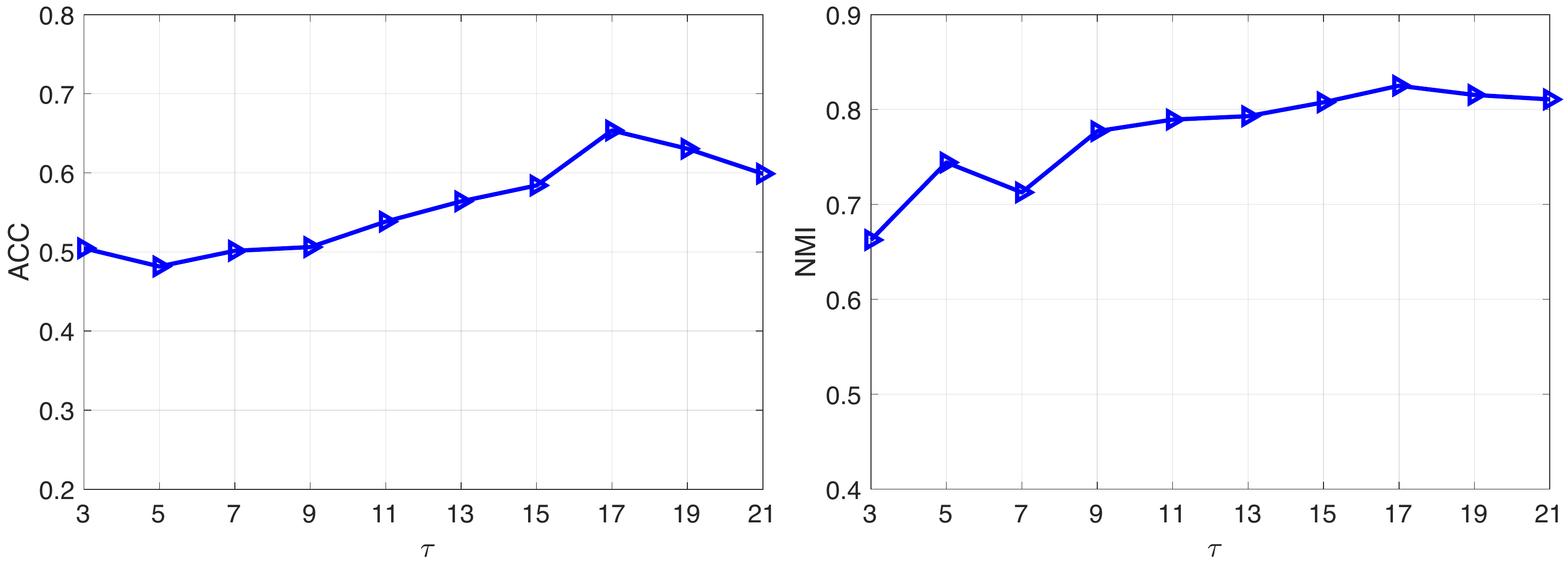}\vspace{-0.15cm}
		\caption{\footnotesize The segmentation performance in terms of NMI and ACC when using different numbers of sequential neighbors $\tau$.}\vspace{-0.45cm}
		\label{fig06}
	\end{center}
\end{figure}

\begin{figure}
	\begin{center}

		\includegraphics[width=0.49\textwidth]{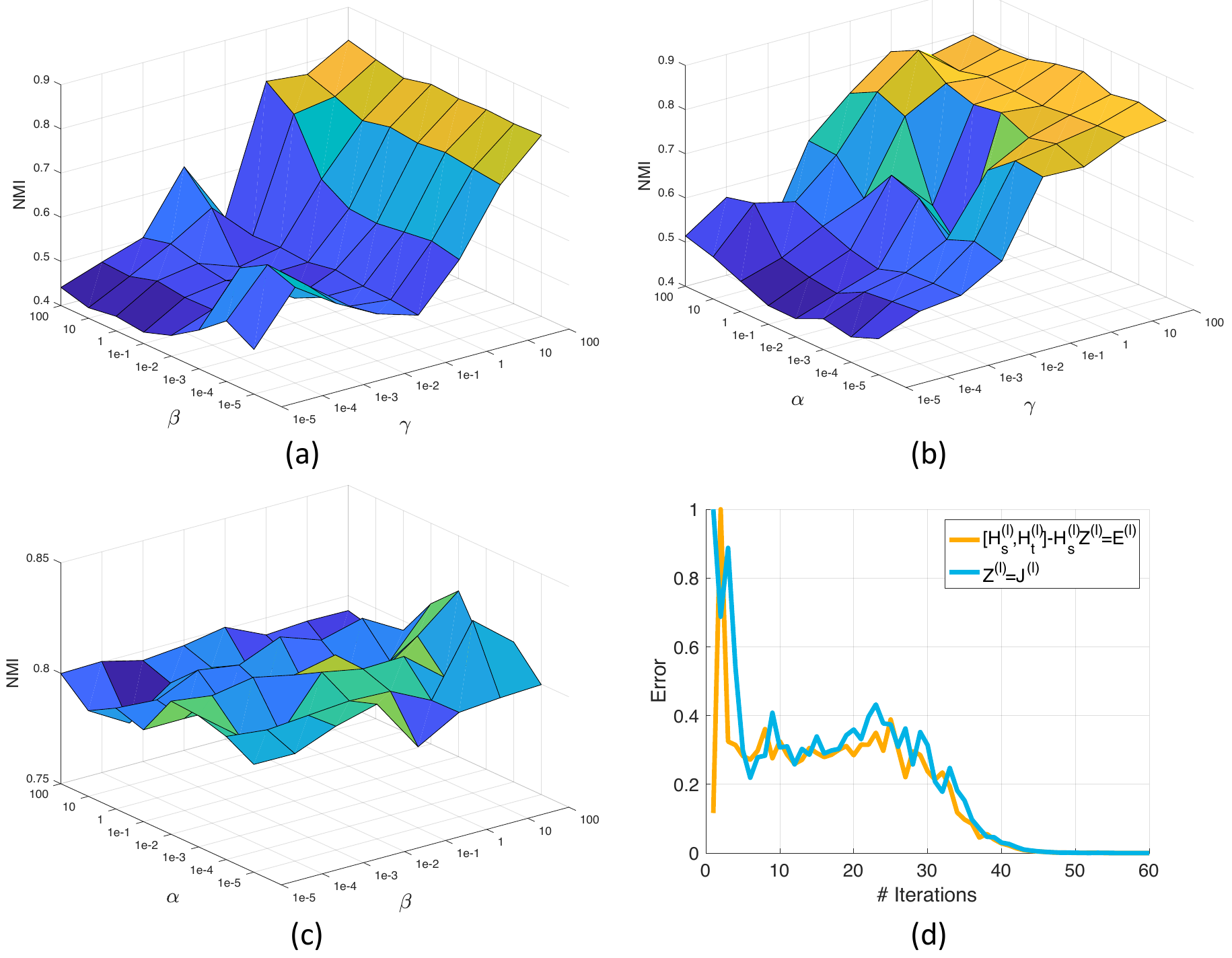}
		\caption{\footnotesize Parameter sensitivity study and convergence analysis on the Keck dataset using HOG features: (a) Sensitivity analysis for parameters $\beta$ and $\gamma$, (b) Sensitivity analysis for parameters $\alpha$ and $\gamma$, (c) Sensitivity analysis for parameters $\alpha$ and $\beta$, and (d) Convergence curves.}
		\label{fig07}
	\end{center}\vspace{-0.5cm}
\end{figure}

In our model, $\tau$ denotes the number of sequential neighbors in the temporal correlation preservation term. To investigate the effect of this parameter, we test our model on the MAD dataset with Keck as the source for different values of $\tau$. Fig.~\ref{fig06} shows the clustering performance for various values within $\{3, 5, \dots, 21\}$.
From the results, it can be observed that our model obtains relatively better performance when $\tau \in [15, 21]$.
In addition, three key regularization parameters, \emph{i.e.}, $\alpha$, $\beta$ and $\gamma$, need to be manually tuned. To investigate the effects of the three parameters on the model output, we fix the value of one parameter and change the other two. The experimental results on the MAD dataset are shown in Fig.~\ref{fig07} (a)(b)(c). From the results, it can be seen that our proposed method obtains much better NMI performance when $\alpha\in[0.001,10]$, $\beta\in[1,100]$, and $\gamma\in[10,100]$. Moreover, the experimental results also indicate that each term in our framework is useful for boosting the segmentation results.

\begin{figure}
	\begin{center}

		\includegraphics[width=0.49\textwidth]{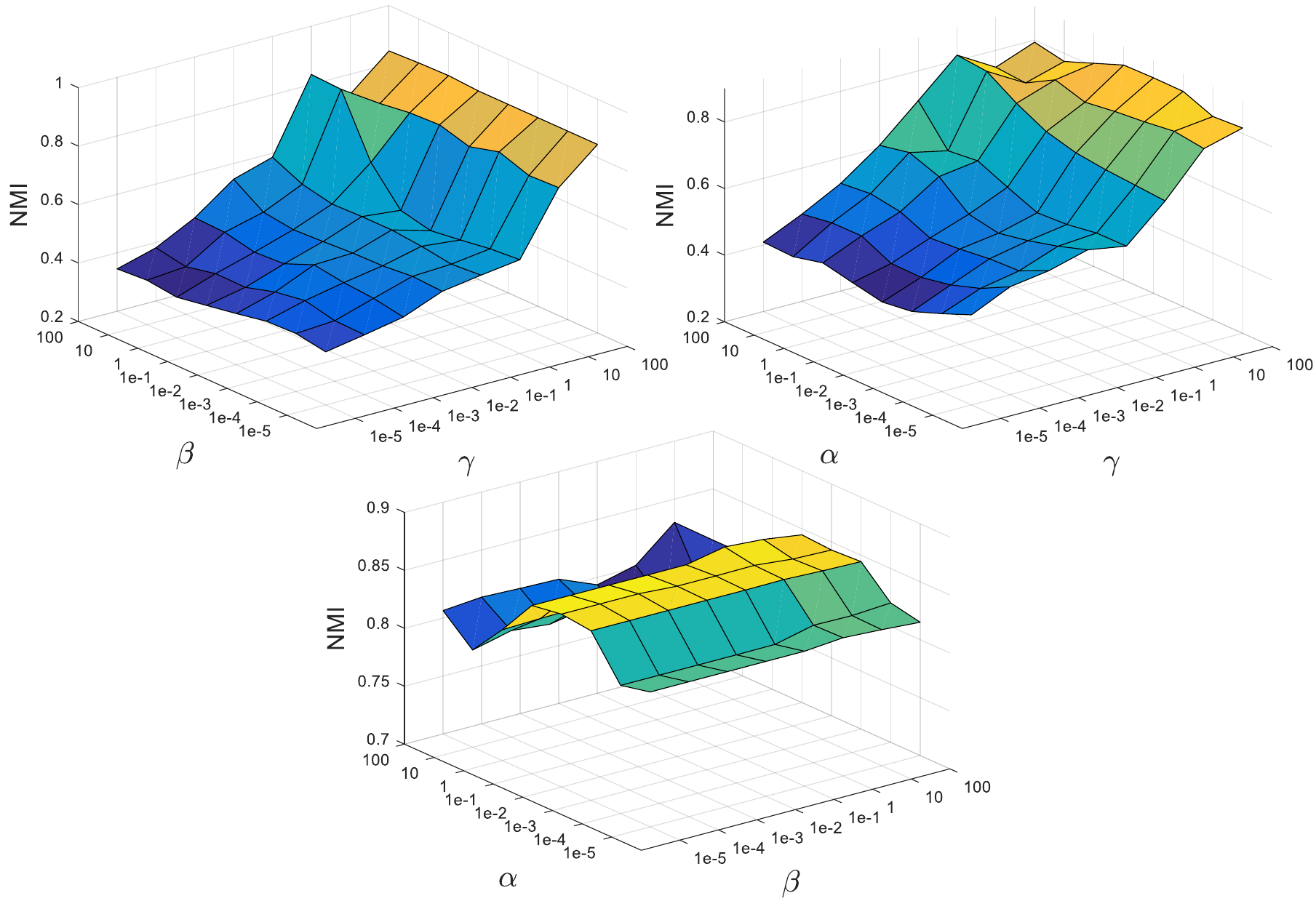}
		\caption{\footnotesize Parameter sensitivity study on MAD: (a) Sensitivity analysis for parameters $\beta$ and $\gamma$, (b) Sensitivity analysis for parameters $\alpha$ and $\gamma$, and (c) Sensitivity analysis for parameters $\alpha$ and $\beta$.}
		\label{fig08}
	\end{center}\vspace{-0.5cm}
\end{figure}

To validate the influence of three parameters for deep features, we conduct an experiment for parameter sensitivity study on MAD using Keck as the source. As shown in Fig.~\ref{fig08}, our method obtains better performance when $\alpha\in[0.001,1]$, $\beta\in[0.001,10]$, and $\gamma\in[10,100]$. Overall, three regularization parameters can be tuned to obtain good performance with different settings.
Besides, we can follow the suggestions when applying our model to new datasets.

\subsubsection{Convergence Analysis}
We compute the errors (\emph{i.e.}, {\small$\|[\mathbf{H}_s^{(l)},\mathbf{H}_t^{(l)}]-\mathbf{H}_s^{(l)}\mathbf{Z}^{(l)}-\mathbf{E}^{(l)}\|_{\infty}$} and {\small$\|\mathbf{W}^{(l)}-\mathbf{J}^{(l)}\|_{\infty}$}) to demonstrate the convergence of our optimization algorithm. We report the mean values of different layers, and the convergence curves on the MAD dataset (Keck as source) are presented in Fig.~\ref{fig07} (d). Note that, for better presentation, the errors are normalized into the range $[0, 1]$. As can be observed, our model converges within about 50 iterations.

\begin{figure}
	\begin{center}
		\includegraphics[width=0.48\textwidth]{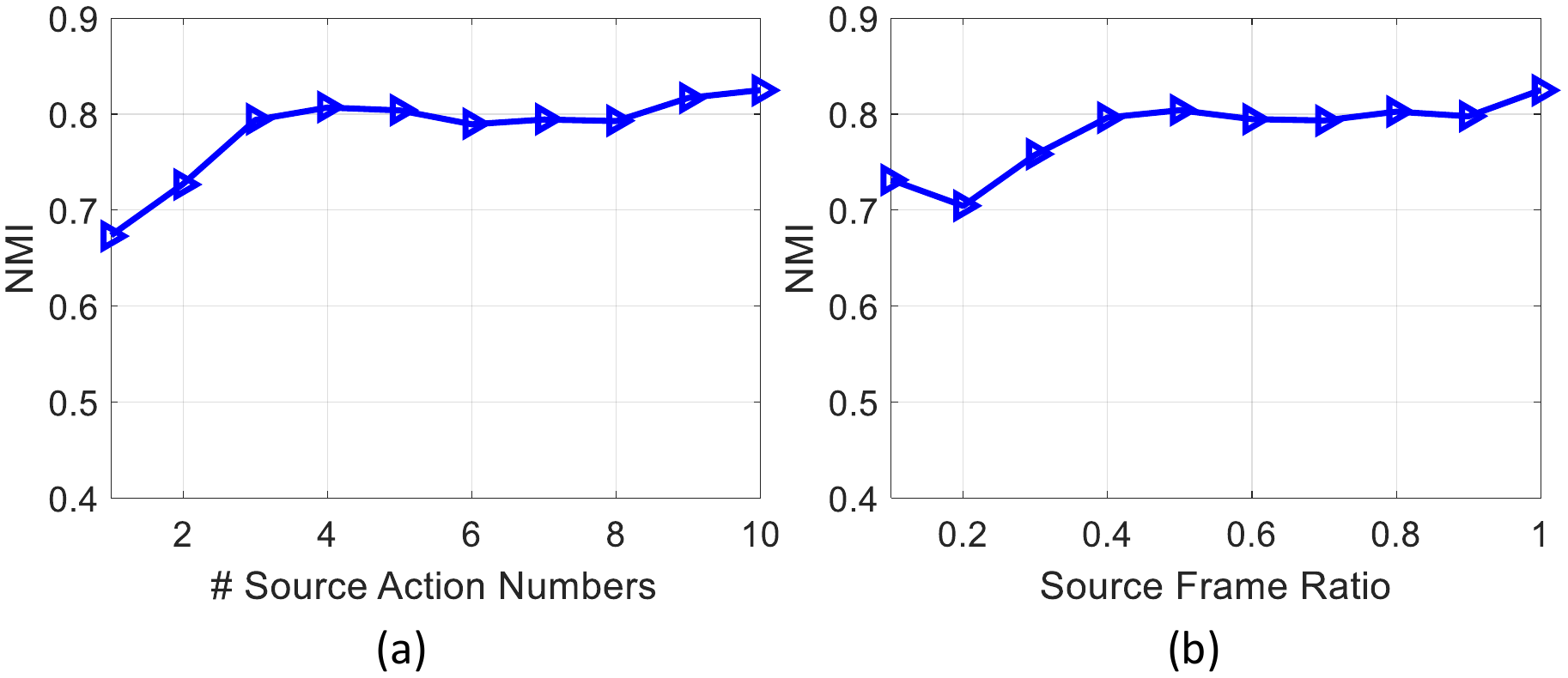}
		\caption{\footnotesize Segmentation results based on (a) using different action numbers and (b) using different frame ratios.}
		\label{fig09}
	\end{center}\vspace{-0.25cm}
\end{figure}

\subsubsection{Source Data Analysis}

To evaluate the effectiveness of the source information for boosting the segmentation performance, we carry out an experiment on the MAD dataset (with Keck as the source) in which different numbers of actions are used. The results are shown in Fig.~\ref{fig09} (a). As can be observed, the performance increases in general when the number of actions increases. This suggests that diverse source data is helpful for boosting the segmentation performance in the target domain. The main reason could be that more actions in the source data can transfer more useful knowledge to ensure that our model learns the distinctive representations of the target data. In addition, we also test the effect of using different numbers of each action. To achieve this, we utilize the frames with different ratios (\emph{i.e.}, $0.1,0.2,\cdots,1$) in each source action, while keeping the total number of actions fixed. We evaluate the performance on the MAD dataset, and the comparison results are shown in Fig.~\ref{fig09} (b). From the results, it can be observed that the performance of our model increases when the ratio of frames increases. This demonstrates that using more frames of each action in the source data can transfer useful knowledge to learn distinctive and effective representations of the target data. Thus, the effectiveness of the source data in improving the segmentation performance can be well validated.

It can be also noted that the results of our method with different source datasets are similar. If it needs to select a good source dataset to further improve the segmentation performance on the target data, we suggest selecting one source dataset that has a similar scenario to the target dataset. For example, we can obtain the best performance on MAD by using Keck as the source, since all these two datasets are under an indoor scenario.

Moreover, we have conducted an experiment by using two source datasets, and the comparison results are shown in Table~\ref{tab04}. From the results, comparing using Weiz and Weiz $\&$ Keck as the sources, our method obtains similar performance in the two settings. Comparing using UT and UT $\&$ Weiz as the sources, our method gets a slightly better performance when using two datasets as sources than using single. Overall, our model can obtain similar and slightly better performance via using two datasets as the source than a single source, but the computation cost will increase when using more datasets. Thus, to balance the performance and computation cost, we just use one dataset as the source in this study.
Additionally, our model is able to flexibly explore the complementarity among multi-level feature spaces, which can also be extended to other multi-modal learning tasks. In future work, we believe that the data size can be reduced by performing sampling techniques or replacing the original data set with a small number of points \cite{chen2011large,tremblay2020}, and binary representations \cite{zhang2018binary} can also be considered to accelerate the computation speed. Moreover, another potential strategy is finding the most diverse and representative frames based on clustering methods, and then we can only use these frames as the source data for reducing the complexity.

\renewcommand\arraystretch{1.0}
\begin{table}[h]
	\centering
	\scriptsize
	\caption{Comparisons on MAD when using one and two datasets as the source.}\vspace{-0.15cm}
	\begin{tabular}{p{2.5cm}||p{1.0cm}|p{1.0cm}} 
		
		\toprule
		& ACC       & NMI \\
     	
     	\midrule
        Weiz (source)	        & 0.6392	& 0.8238 \\
        Weiz + Keck (source)	& 0.6345	& 0.8325 \\

        \midrule
        UT (source)	            & 0.6371	& 0.8238 \\
        UT + Weiz (source)	    & 0.6440	& 0.8289 \\

		\bottomrule
	\end{tabular}
	\label{tab04}\vspace{-0.05cm}
\end{table}

\subsubsection{Dictionary Dimensionality Analysis}

\begin{figure}
	\begin{center}
		\includegraphics[width=0.5\textwidth]{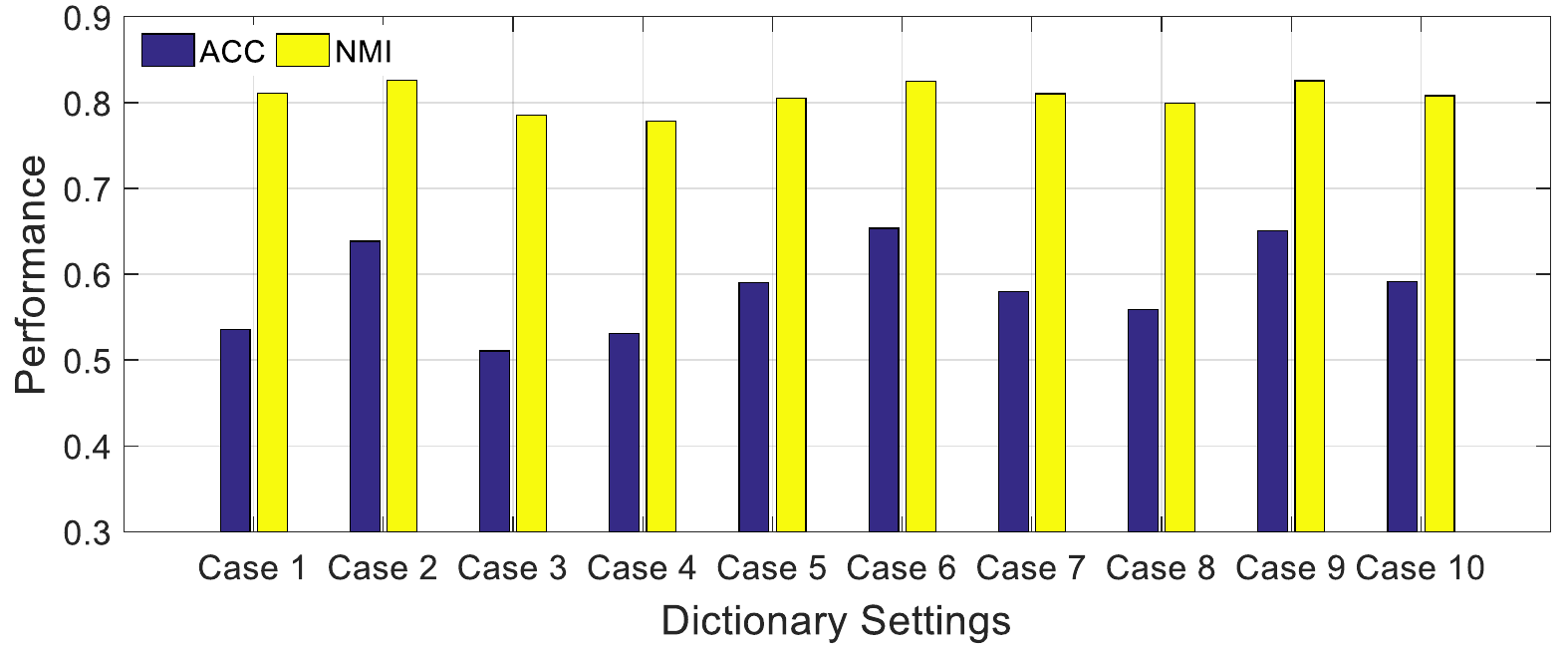}\vspace{-0.2cm}
		\caption{\footnotesize Segmentation results on MAD (Keck as source) via using different numbers of layers and dimensionalities of dictionary atoms.}
		\label{fig10}
	\end{center}\vspace{-0.2cm}
\end{figure}

In the proposed model, we adopt a multi-layer deep matrix factorization structure, thus we need to set different dimensions for dictionary atoms at each layer. To investigate the effects of dimensionality of the dictionary atoms, we conduct a comparison experiment using different dimension sets. Specifically, we set ten cases in the dimension set,
\ie, $\{128$-$32$; $128$-$16$; $64$-$32$; $64$-$16$; $32$-$16$; $128$-$64$-$16$; $128$-$32$-$16$; $64$-$32$-$16$; $128$-$64$-$32$-$16$;
$128$-$64$-$32$-$16$-$8\}$.
Fig.~\ref{fig10} shows the segmentation results obtained based on using different numbers of layers and dimensionalities of the dictionary atoms on MAD using Keck as the source data. From the results, the ACC results vary when using different dimensionalities of the dictionary atoms, while the NMI results are robust to dimensionalities of the dictionary atoms. Thus, in this study, we adopt a three-layer deep NMF structure (\ie, $128$-$64$-$16$) for all comparisons as the default setting.

\subsubsection{Ablation Study}

\begin{figure}[h]
	\begin{center}

		\begin{tikzpicture} 
		\begin{axis}[
		width=0.34\textwidth,
		height=0.12\textheight,
		ymin=0.65,
		ymax=0.85,
		ytick={0.65,0.75,0.85},
		ybar,
		enlargelimits=0.2,
		legend style={
			at={(1.25,0.52)},
			anchor=north,
			legend columns=1},
		legend cell align=left,
		ylabel={NMI},
		ylabel style={font=\footnotesize},
		yticklabel style = {font=\scriptsize},
		xticklabel style = {font=\scriptsize},
		legend style={draw=none},
		legend style={font=\scriptsize\sffamily},
		xtick=data,
		    axis x line*=bottom,
		axis y line*=left,
		bar width=9.5pt,
		xticklabels={Keck,MAD,UT},
		]
		
		\addplot+[error bars/.cd, 
		y dir=both,y explicit]
		[color=black!90, very thick, fill=green!120, very thick]
		coordinates {(1,0.8189)
			(2,0.7707)
			(3,0.7966) };
		
		\addplot+[error bars/.cd, 
		y dir=both,y explicit]
		[color=black!90, fill=blue!120, very thick]
		coordinates {(1,0.7930)
			(2,0.7921)
			(3,0.7991) 	};
		
		\addplot+[error bars/.cd,  %
		y dir=both,y explicit]
		[color=black!90, fill=red!120, very thick]
		coordinates {(1,0.8601)
			(2,0.8375)
			(3,0.8590) };

		\legend{First layer,Last layer,Layers fusion}
		\end{axis} 
		\end{tikzpicture}\vspace{-0.1cm}
		{Source Data}
		\caption{\footnotesize Performance comparison (NMI) when using representations from different layers or multi-layer fusion.}\label{fig11}
	\end{center}

\end{figure}
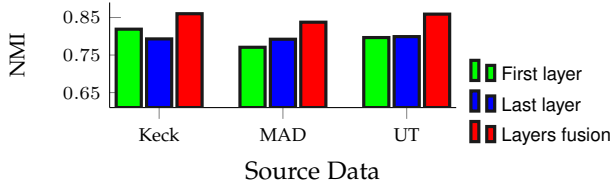

\textbf{Effectiveness of Multi-level Representations}: To validate the effectiveness of fusing multi-level subspace representations from different feature spaces, we conduct an experiment to test the performance of our model when using the representations from the first layer, last layer, and multiple fused layers on the Weiz dataset. The results are shown in Fig.~\ref{fig11}. We can see that our fusion strategy obtains much better performance than conducting subspace learning only on the representations from the first layer or last layer. This indicates the effectiveness of our model, which fuses the multi-level subspace representations for transfer learning.

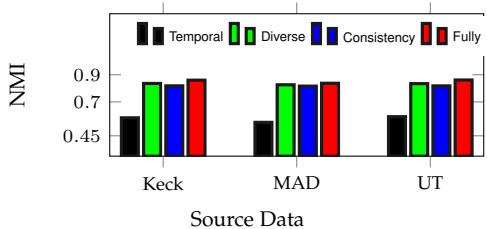
\begin{figure}[h]
	\begin{center}
		\captionsetup{font={small}}

		\begin{tikzpicture} 
		\begin{axis}[
		width=0.36\textwidth,
		height=0.14\textheight,
		ymin=0.45,
		ymax=1.2,
		ytick={0.45,0.7,0.9},
		ybar,
		enlargelimits=0.2,
 legend style={
    at={(0.05,0.7)},
     font=\tiny,
      anchor=south west,
      legend columns=-1},
      legend cell align=left,
      legend style={draw=none},
    ylabel={NMI},
    ylabel style={font=\footnotesize},
    yticklabel style = {font=\scriptsize},
    xticklabel style = {font=\scriptsize},
    legend style={font=\tiny\sffamily},
    xtick=data,
    bar width=6.4pt,
    xticklabels={Keck,MAD,UT},
		]
		
		\addplot+[error bars/.cd, 
		y dir=both,y explicit]
		[color=black!90, very thick, fill=black!120, very thick]
		coordinates {(1,0.5826)
			(2,0.5492)
			(3,0.5910) };

		\addplot+[error bars/.cd,
		y dir=both,y explicit]
		[color=black!90, very thick, fill=green!120, very thick]
		coordinates {(1,0.8360)
			(2,0.8262)
			(3,0.8346) };

		\addplot+[error bars/.cd, 
		y dir=both,y explicit]
		[color=black!90, fill=blue!120, very thick]
		coordinates {(1,0.8183)
			(2,0.8162)
			(3,0.8181) 	};
		
		\addplot+[error bars/.cd,  
		y dir=both,y explicit]
		[color=black!90, fill=red!120, very thick]
		coordinates {(1,0.8601)
			(2,0.8375)
			(3,0.8616) };

		\legend{Temporal, Diverse,Consistency,Fully}
		\end{axis} 
		\end{tikzpicture}\vspace{0.05cm}\\
		\footnotesize{Source Data}
		\caption{\footnotesize Results comparison of different key components.}\label{fig12}
	
	\end{center}\vspace{-0.9cm}
	~~~~~~~~~~~
\end{figure}

\textbf{Effectiveness of Different Key Components}: To validate the effectiveness of different key components (\ie, the temporal correlation preservation term, diversity across multi-level representation term, and multi-mutual consistency learning term), we test our model when if moving one component and keeping the other two, on the Weiz dataset. Fig.~\ref{fig12} shows the comparison results of different key components. From these results, it can be observed that the temporal correlation preservation term plays an important role in human motion segmentation, which effectively preserves temporal correlations among consecutive frames. Moreover, we can see that the diversity constraint and consistency learning are also helpful for boosting the segmentation performance. Thus, these three key components work together to improve the model performance.

\section{Conclusion}
\label{Conclusion}
We have proposed a novel CDMS framework for human motion segmentation. Our model first factorizes the original features of the source and target data into implicit multi-layer feature spaces, in which we carry out transfer subspace learning on different layers to capture multi-level structural information. A multi-mutual consistency learning strategy is carried out to reduce the distribution gap between the source and target domains. Further, we introduce a novel constraint term based on the HSIC to strengthen the diversity of multi-level subspace representations, which enables the complementarity of multi-level representations to be explored in order to boost the transfer learning performance. Moreover, we develop an enhanced graph regularizer term to preserve the temporal correlations. Finally, we obtain the optimized affinity matrix by fusing the multi-level subspace representation coefficients. Extensive experimental results on benchmark datasets demonstrate the effectiveness of the proposed model against several state-of-the-art human motion segmentation methods.

%\section*{Acknowledgments}
%This work was supported in part by the National Natural Science Foundation of China (Nos: 62172228, 61973162), NSF of Jiangsu Province (No: BZ2021013), the Fundamental Research Funds for the Central Universities (Nos: 30920032202, 30921013114), and National Key R\&D Program of China (No: 2021YFA1001100).

\bibliographystyle{IEEEbib}
% \bibliography{egbib}

\vfill

% or


%\appendices

%\section{}

%\ifCLASSOPTIONcompsoc
%\else
%\fi


\ifCLASSOPTIONcaptionsoff
  \newpage
\fi

\end{document}